\documentclass[10pt,twocolumn,letterpaper]{article}

\usepackage[pagenumbers]{cvpr} 
\usepackage{amssymb}
\usepackage{amsmath}
\usepackage{pifont}  
\usepackage{subcaption} 
\usepackage{caption}
\usepackage{booktabs,multirow}
\usepackage[accsupp]{axessibility}
\usepackage{tikz}
\usepackage[ruled,vlined,linesnumbered]{algorithm2e}
\DontPrintSemicolon
\SetKwInput{KwInput}{Input}
\SetKwInput{KwOutput}{Output}
\DeclareMathOperator*{\argmin}{arg\,min}

\usetikzlibrary{arrows.meta,positioning,shapes.geometric,fit,calc,backgrounds,decorations.pathreplacing}

\definecolor{cvprblue}{rgb}{0.21,0.49,0.74}
\usepackage[pagebackref,breaklinks,colorlinks,allcolors=cvprblue]{hyperref}

\newcommand{\increment}{\mathcal{I}}

\def\paperID{****} 
\def\confName{CVPR}
\def\confYear{2026}

\title{Learnability-Guided Diffusion for Dataset Distillation}

\author{Jeffrey A. Chan-Santiago\\
Institute of Artificial Intelligence \\
University of Central Florida\\
{\tt\small jeffrey.chansantiago@ucf.edu}
\and
Mubarak Shah\\
Institute of Artificial Intelligence \\
University of Central Florida\\
{\tt\small shah@crcv.ucf.edu}
}

\begin{document}
\maketitle
\begin{abstract}
Training machine learning models on massive datasets is expensive and time-consuming. Dataset distillation addresses this by creating a small synthetic dataset that achieves the same performance as the full dataset. Recent methods use diffusion models to generate distilled data, either by promoting diversity or matching training gradients. However, existing approaches produce redundant training signals, where samples convey overlapping information. Empirically, disjoint subsets of distilled datasets capture 80--90\% overlapping signals. This redundancy stems from optimizing visual diversity or average training dynamics without accounting for similarity across samples, leading to datasets where multiple samples share similar information rather than complementary knowledge.
We propose learnability-driven dataset distillation, which constructs synthetic datasets incrementally through successive stages. Starting from a small set, we train a model and generate new samples guided by learnability scores that identify what the current model can learn from, creating an adaptive curriculum.
We introduce Learnability-Guided Diffusion (LGD), which balances training utility for the current model with validity under a reference model to generate curriculum-aligned samples. Our approach reduces redundancy by 39.1\%, promotes specialization across training stages, and achieves state-of-the-art results on ImageNet-1K (60.1\%), ImageNette (87.2\%), and ImageWoof (72.9\%). Our code is available on our project page\footnote{\url{https://jachansantiago.github.io/learnability-guided-distillation/}}.
\end{abstract}
\vspace{-1em}

\section{Introduction}
\label{sec:intro}

Dataset distillation has attracted broad interest for its promise to dramatically reduce training data requirements without sacrificing model performance. The goal is to synthesize a small surrogate dataset $D_\mathcal{S}$ from a large target dataset $D_\mathcal{T}$ such that models trained on $D_\mathcal{S}$ achieve comparable accuracy to those trained on $D_\mathcal{T}$. Guo et al. \cite{guo2024datm} demonstrated near-lossless accuracy on small-scale benchmarks: on CIFAR-10 and CIFAR-100 \cite{krizhevsky2009learning}, distilled datasets with only 100 images per class (IPC) match the full 5,000 IPC datasets---a 50$\times$ compression using $\sim 2\%$ of the data. However, scaling to larger, high-resolution datasets like ImageNet remains challenging.

\begin{figure}[t]
    \centering
    \includegraphics[page=1,width=0.9\linewidth,trim=184mm 96mm 15mm 9mm,clip]{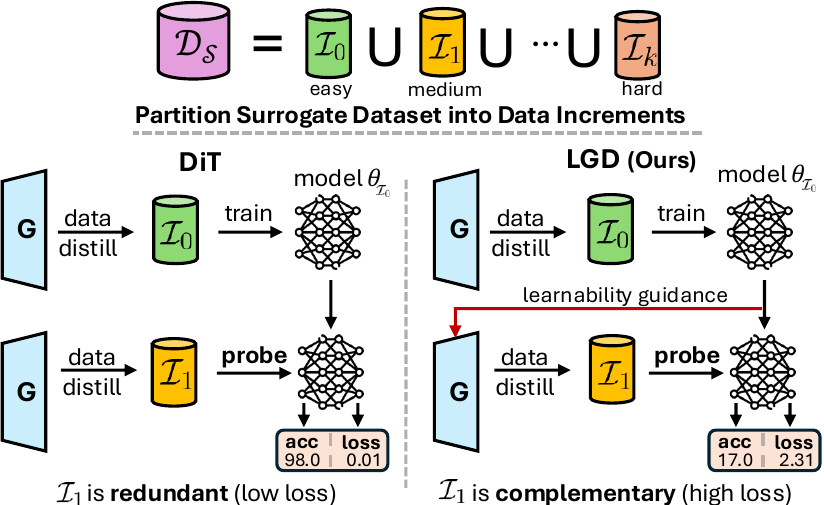}
    \caption{\textbf{Learnability-Guided Dataset Distillation.} We partition the distilled dataset $\mathcal{D}_S$ into increments $\{\mathcal{I}_0, \mathcal{I}_1, \ldots, \mathcal{I}_k\}$ ({\bf Top}). \textbf{Bottom Left (DiT):} Standard distillation generates increments independently, producing redundant samples---a model trained on $\mathcal{I}_0$ achieves 98.0\% accuracy on $\mathcal{I}_1$, indicating no new information. \textbf{Bottom Right (LGD):} We condition the next increment on the model parameters $\theta_{\mathcal{I}_0}$ to guide synthesis toward samples that complement $\mathcal{I}_0$. The resulting increment $\mathcal{I}_1$ achieves only 17.0\% accuracy when evaluated by the prior model, indicating it introduces substantial new learning signal.}
    \label{fig:teaser}
    \vspace{-1.3em}
\end{figure}

Dataset distillation methods traditionally synthesize $D_\mathcal{S}$ through bi-level optimization that matches training trajectories \cite{cazenavette2022dataset, guo2024datm, cui2023scaling, zhao2021datasetcondensation, shin2023lcmat}—optimizing synthetic data such that model parameters trained on $D_\mathcal{S}$ evolve similarly to those trained on $D_\mathcal{T}$.
While effective on small datasets, this becomes computationally intractable for high-resolution images. To address scalability, generative dataset distillation \cite{zhang2023dc, su2024d4m, gu2024efficient, chen2025igd, chan-santiago2025mgd3} leverages pretrained generative models to synthesize distilled datasets at much lower cost. Recent work shows trajectory matching remains valuable: influence-guided generation \cite{chen2025igd} steers samples whose gradients align with those from training on $D_{\mathcal{T}}$.

\begin{figure}[tbh]
    \centering
    \includegraphics[width=\linewidth]{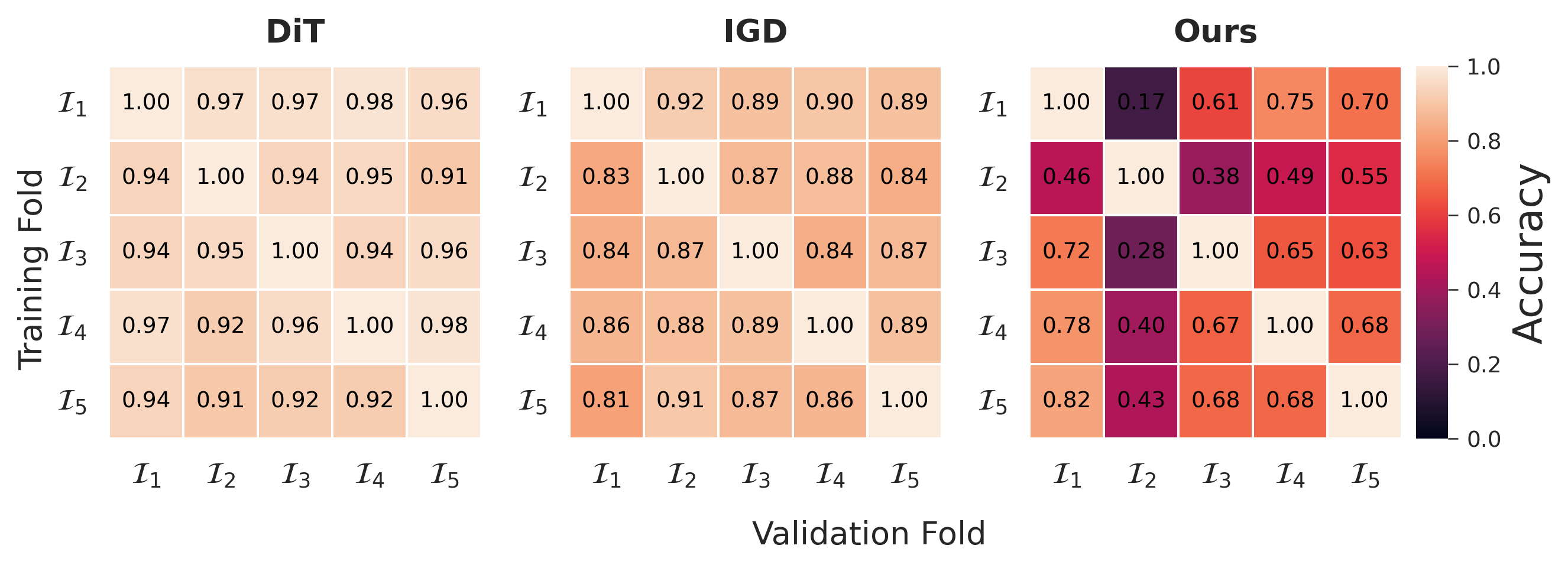}
    \vspace{-2em}
        \caption{
\textbf{Cross-validation across distilled data increments} (\(\mathcal{I}_1\!-\!\mathcal{I}_5\)) \textbf{for IPC 50 on ImageNette.}
Each heatmap shows accuracy when training on one increment (rows) and evaluating on another (columns).
DiT\cite{peebles2023scalable} and IGD \cite{chen2025igd} exhibit high cross-increment accuracy due to overlapping information, while LGD yields lower off-diagonal scores, indicating more \textbf{complementary and diverse} increments.}
    \label{fig:crossval}
\vspace{-1.5em}
\end{figure}

However, this has a fundamental limitation: optimizing all samples toward the \emph{average} training trajectory causes convergence to similar gradient profiles rather than specialization across training phases. Model training naturally progresses through stages: early training benefits from samples with strong gradients for coarse features, while late training requires small, refined gradients for fine-grained details. A sample cannot satisfy both—optimizing for the average produces medium-strength gradients throughout, useful at no specific stage. We confirm this empirically: partitioning a 50 IPC dataset into five disjoint 10 IPC subsets, any subset captures 80--90\% of the training signal from others (\cref{fig:teaser,fig:crossval}), demonstrating redundancy.

To address this, we propose learnability-driven distillation that builds the synthetic dataset incrementally. Starting from a small initial dataset (e.g., IPC = 10), we train a model to convergence, then synthesize new samples guided by learnability scores—generating samples that complement rather than replicate existing data. This incremental approach reduces redundancy by conditioning each stage on the model's evolving learning frontier.
This reframes distillation as sequential learning: \emph{given a distilled dataset and a model trained on it, generate additional samples that maximize marginal learning gains}. This approach offers a principled way to reduce redundancy and establish a foundation for methods that exploit staged learning dynamics.

\noindent\textbf{Contributions.} Our main contributions are:

\begin{itemize}
    \item \emph{Learnability-driven incremental framework}. (\cref{sec:incremental_formulation}) We construct distilled datasets stage-by-stage, conditioning each increment on the current model's learnability to generate complementary rather than redundant training signals.
    
    \item \emph{Learnability-guided synthesis.} (\cref{sec:learnability_guidance})  We propose Learnability-Guided Diffusion (LGD) that conditions generation on the current model state to synthesize samples that complement existing data, integrated into diffusion sampling to automatically generate informative increments.

    \item \emph{Redundancy analysis.} (\cref{sec:redundancy}) Our incremental framework enables quantifying information overlap in distilled datasets, revealing 80--90\% redundancy in existing methods.
    
    \item \emph{Improved sample efficiency.} (\cref{sec:static,sec:redundancy}) We achieve state-of-the-art or competitive results on ImageNet-1K (60.1\%), ImageNette (82.6--87.2\%), and ImageWoof (53.9--72.9\%), while reducing redundancy by 39.1\%.
\end{itemize}

\section{Related Work}
\label{sec:related_work}
\textbf{Dataset Distillation.}
Early methods synthesize compact datasets through optimization that matches training dynamics using gradient matching~\cite{zhao2021datasetcondensation, shin2023lcmat}, distribution matching~\cite{zhao2023distribution, wang2022cafe, zhao2023idm, sajedi2023datadam}, or trajectory matching~\cite{cazenavette2022dataset, guo2024datm, cui2023scaling}. While effective on small benchmarks, these pixel-optimization approaches struggle with high-resolution datasets. Decoupled methods~\cite{yin2023sre2l, shao2024edc} improve scalability through sequential stages, enhanced by multi-architecture training~\cite{shao2024gvbsm}, patch-based composition~\cite{sun2024rded}, and curriculum strategies~\cite{ma2025cudd}. However, pixel-space constraints limit sample diversity, motivating generative approaches.

\textbf{Generative Dataset Distillation.}
Recent works \cite{zhao2022synthesizing, zhang2023dc, gu2024efficient, su2024d4m, chen2025igd, chan-santiago2025mgd3} leverage pretrained generative models to overcome pixel-space limitations. Diffusion-based methods offer realistic generation: Minimax Diffusion~\cite{gu2024efficient} balances diversity and representativeness through diffusion fine-tuning, MGD$^3$~\cite{chan-santiago2025mgd3} and D$^4$M~\cite{su2024d4m} conditions on feature-space modes for broader coverage, and Influence-Guided Diffusion (IGD)~\cite{chen2025igd} guides generation to match training gradients from the full dataset. However, these methods synthesize all samples uniformly, causing convergence to similar gradient profiles. 
Yet these approaches suffer from inherent redundancy because they optimize for visual diversity or average training trajectories without accounting for training signal similarity across samples. This produces datasets where multiple samples teach the model similar information rather than providing complementary knowledge across training stages.
Our work addresses this through learnability-driven synthesis: conditioning each increment on the current model state to generate complementary rather than redundant samples.

\textbf{Curriculum Learning and Data Curation.}
Our approach connects to curriculum learning, where training difficulty evolves with model competence~\cite{mirzasoleiman2020coresets, curriculum2009bengio}. Recent methods~\cite{He_2024_CVPR, zhang2024spanning, evans2024data, zhou2025scale} identify learnable samples by tracking training signals, filtering samples that are already mastered or too difficult. Feedback-driven synthesis~\cite{askari-hemmat2025improving, askari-hemmat2024feedbackguided} generates harder examples tailored to model weaknesses. 
We extend this to dataset distillation by iteratively synthesizing data increments 
that maximize marginal learning gains, transforming distillation into a 
learnability-driven curriculum that adapts to evolving training needs.

\section{Background}
\label{sec:background}

\textbf{Dataset Distillation.} 
Dataset distillation aims to create a small synthetic dataset $\mathcal{D} = \{(\tilde{x}_i, \tilde{y}_i)\}_{i=1}^{M}$ 
that captures the essential information from a large training dataset 
$\mathcal{T} = \{(x_i, y_i)\}_{i=1}^{N_{\mathcal{T}}}$, where $M \ll N_\mathcal{T}$. 
The key requirement is that models trained on the small dataset $\mathcal{D}$ 
should perform nearly as well as models trained on the full dataset $\mathcal{T}$. 
We denote these models as $\theta_{\mathcal{D}}$ and $\theta_{\mathcal{T}}$ respectively, 
with the goal that $\mathcal{A}(\theta_{\mathcal{D}}) \approx \mathcal{A}(\theta_{\mathcal{T}})$, 
where $\mathcal{A}(\cdot)$ measures test accuracy. 
The budget for distillation is specified in images per class (IPC).
Modern approaches~\cite{gu2024efficient,zhao2022synthesizing, zhang2023dc, su2024d4m, chan-santiago2025mgd3, chen2025igd} use generative models to synthesize $\mathcal{D}$ by matching the learning behavior on real data:
\begin{equation}
\label{eq:distillation_objective}
    \min_{\mathcal{D}} \left\| \mathbb{E}_{x \sim P_{\text{data}}}\big[ \ell (\theta_{\mathcal{T}}(x), y) \big] - 
    \mathbb{E}_{x \sim P_{\text{data}}} \big[ \ell (\theta_{\mathcal{D}}(x), y) \big] \right\|
\end{equation}
where $\ell$ is a loss function and $P_{\text{data}}$ is the data distribution. 
This generative approach can create entirely new samples rather than just selecting 
from existing data, enabling greater flexibility and diversity.

\textbf{Diffusion Models.}
Diffusion models~\cite{ho2020denoising} generate images through a two-stage process. 
First, the \emph{forward process} gradually adds noise to real images over $T$ steps. 
Starting from a clean image $x_0$, each step adds a small amount of Gaussian noise:
$q(x_t | x_{t-1}) = \mathcal{N}(\sqrt{1-\beta_t} x_{t-1}, \beta_t \mathbf{I})$, 
where $\beta_t$ controls how much noise to add at step $t$. 
We can jump directly to any noisy version using 
$x_t = \sqrt{\bar{\alpha}_t} x_0 + \sqrt{1-\bar{\alpha}_t} \epsilon$, 
where $\epsilon \sim \mathcal{N}(0, \mathbf{I})$ and $\bar{\alpha}_t = \prod_{s=1}^t (1-\beta_s)$.

Second, the \emph{reverse process} learns to remove noise step-by-step. A neural network $\epsilon_\phi(x_t, t)$ predicts the noise at each step, and we use this to compute the mean of the reverse distribution:
\begin{equation}
\label{eq:reverse_process}
    \mu_\phi(x_t) = \frac{1}{\sqrt{1-\beta_t}} \left(x_t - 
    \frac{\beta_t}{\sqrt{1-\bar{\alpha}_t}} \epsilon_\phi(x_t, t) \right)
\end{equation}
The denoised sample is then drawn as $x_{t-1} = \mu_\phi(x_t) + \sigma_t \mathbf{z}$, where $\mathbf{z} \sim \mathcal{N}(0, \mathbf{I})$ adds controlled randomness.
Starting from pure noise, we repeatedly apply this denoising step to generate new images. 
For class-conditional generation, we provide the class label as input: $\epsilon_\phi(x_t, t, c)$.

\textbf{Guided Sampling.}
We can steer diffusion models to generate images with specific properties by adding guidance signals during sampling.
Classifier guidance~\cite{dhariwal2021diffusion} adjusts the noise prediction to favor a target class $c$:
\begin{equation}
\label{eq:guidance}
    \tilde{\epsilon}_\phi(x_t, t, c) = \epsilon_\phi(x_t, t, c) + 
    \lambda \nabla_{x_t} \log p(c | x_t)
\end{equation}
where $\lambda$ controls the guidance strength. The gradient term 
$\nabla_{x_t} \log p(c | x_t)$ steers generation toward class $c$. 
Recent work in distillation~\cite{chen2025igd,chan-santiago2025mgd3} uses similar guidance 
based on proxies for training utility—guiding the model to generate samples that will be 
most useful for learning. We build on this idea by making the guidance adaptive 
to the current training state.

\section{Method}
\label{sec:method}
\begin{figure*}[t]
    \centering
    \includegraphics[page=1,width=0.95\linewidth,trim=12mm 96mm 22mm 0mm,clip]{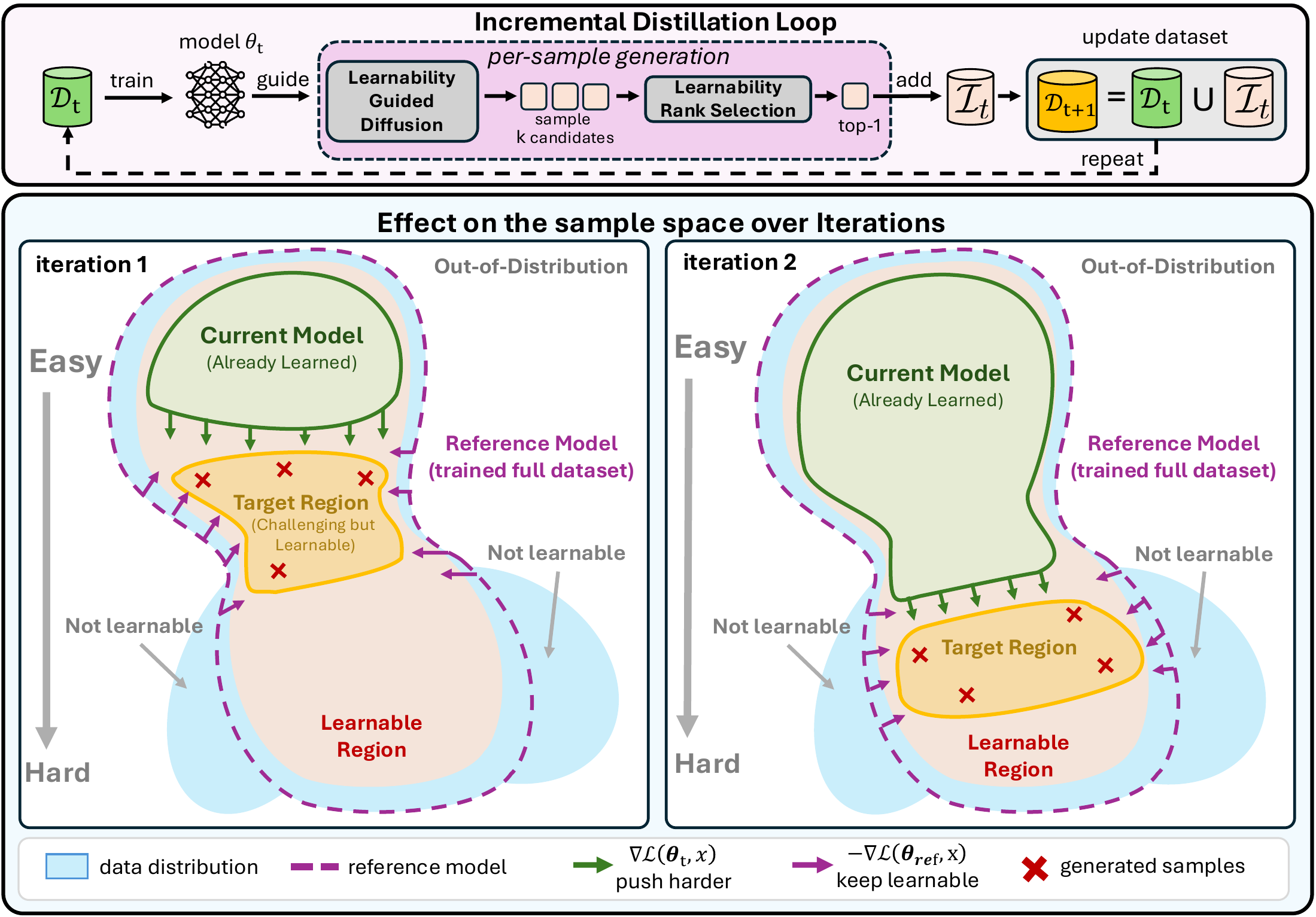}
    \caption{\textbf{Overview of our learnability-guided iterative generation framework.}
(\emph{Top}) Incremental distillation loop: we iteratively train model $\theta_t$ on cumulative dataset $\mathcal{D}_t$, 
generate samples using our learnability guidance, select high-quality samples via learnability ranking, and augment the dataset. (\emph{Bottom}) Effect on sample space: 
The current model $\theta_t$ (green) expands over iterations, while the reference model 
$\theta^*$ (purple, fixed) defines the learnable region. Generated samples (red $\times$) land 
in the learnable gap between boundaries, automatically synthesizing samples that complement 
the current model's learned distribution.}
    \label{fig:overall_method}
\vspace{-1em}
\end{figure*}

Existing dataset distillation methods typically optimize synthetic images to mimic the gradient trajectories or training dynamics obtained from the full dataset. 
In contrast, we propose an \emph{incremental} formulation that builds the distilled dataset stage by stage, where each newly generated increment is optimized to maximize the model’s learning signal given its current knowledge. 
Our key insight is that by aligning the synthesis process with the learner’s evolving state, each synthetic sample can contribute complementary, non-redundant information, leading to a more efficient and adaptive curriculum.

\subsection{Incremental Distillation Formulation}
\label{sec:incremental_formulation}
We formulate learnability-driven distillation as an incremental synthesis problem.
Let $\mathcal{D}$ denote the final distilled dataset of size $M$, partitioned into $K$ disjoint increments $\increment_i = \{(x_j^i, y_j^i)\}_{j=1}^{N_i}$ such that $\sum_{i=1}^{K} N_i = M$ and $\increment_i \cap \increment_j = \emptyset$ for $i \neq j$.
At stage $i$, the model with parameters $\theta_{i-1}$ is trained on the cumulative dataset
{\setlength{\abovedisplayskip}{4pt}%
 \setlength{\belowdisplayskip}{4pt}%
\begin{equation}
\mathcal{D}_i = \bigcup_{k=1}^{i}\increment_k,
\end{equation}
}
resulting in updated parameters $\theta_i$. 
This process repeats until all $K$ increments have been synthesized and incorporated into $\mathcal{D}$.

Given the current model state $\theta_{i-1}$, we seek the next increment $\increment_i$ that maximizes its contribution to learning:
\begin{equation}
\increment_i^* = \arg\max_{\increment}\,\mathcal{L}(\theta_{i-1}, \increment),
\label{eq:increment_objective_naive}
\end{equation}
where $\mathcal{L}(\theta_{i-1}, \increment)$ measures how much the model can learn from $\increment$.
However, without constraints, this process can produce degenerate samples that drift away from meaningful semantics or correct labels.
We therefore regularize the synthesis with a reference model $\theta^*$ trained on the full dataset:
\begin{equation}
\increment_i^* = \arg\max_{\increment}\,[\mathcal{L}(\theta_{i-1}, \increment) - \mathcal{L}(\theta^*, \increment)].
\label{eq:increment_objective}
\end{equation}
The second term acts as a regularizer, penalizing samples that are difficult or misclassified by the reference model.
This objective naturally promotes the generation of examples that are hard for $\theta_{i-1}$ yet still learnable under $\theta^*$, ensuring each increment targets learnable knowledge gaps. 

\textbf{Incremental distillation as synthesis and analysis framework.}
Beyond synthesis, our incremental formulation provides an \emph{analysis 
framework} for diagnosing sample redundancy. By 
partitioning any dataset $\mathcal{D}$ into increments and evaluating 
cross-increment learning dynamics, we can quantify information overlap 
across different distillation methods. For synthesis, each increment is 
conditioned on the evolving model state to maximize complementary information. 
For evaluation, the final dataset $\mathcal{D}$ is compatible with both 
incremental training and standard static protocols, enabling direct comparison 
with prior work. \cref{fig:overall_method} (top) illustrates this incremental
loop: at each stage, we train model $\theta_i$ on the cumulative dataset, 
generate candidates via learnability-guided diffusion, select high-quality 
samples, and augment the dataset for the next iteration.

\subsection{Data Seed Initialization}
\label{sec:data_seed}
Our learnability-driven synthesis requires an initial seed dataset  $\mathcal{D}_1$. By default, we use IGD ~\cite{chen2025igd} distilled images at 10 IPC per class, though  $\mathcal{D}_1$ can also be sampled from pretrained diffusion models or other distilled datasets. This seeded initialization accelerates convergence by starting from a distilled dataset rather than random samples. Ablations on seed choices are in the supplementary material.

\subsection{Learnability-Guided Diffusion Sampling}
\label{sec:learnability_guidance}
During synthesis, we incorporate a learnability criterion derived from \cref{eq:increment_objective} into the diffusion sampling process, guiding the denoising trajectory toward regions with high learnability scores.

\textbf{Learnability Score.}
For a candidate sample $(x, y)$, we define its learnability \cite{mindermann2022prioritized, evans2024data} as
\begin{equation}
\mathcal{S}(x, y) = \mathcal{L}(\theta_{i-1}, x, y) - \omega \cdot \mathcal{L}(\theta^*, x, y),
\label{eq:learnability_score}
\end{equation}
where $\mathcal{L}(\theta, x, y)$ is a prediction loss (e.g., cross-entropy) and $\omega$ controls the reference model regularization strength.
High $\mathcal{S}$ indicates the current model struggles while the reference model does not—the sample is both informative and semantically valid.

\begin{table*}[tbh]
\centering
\caption{\textbf{Comparison across distilled IPC budgets on Nette and Woof evaluated on different network architectures}. Mean$\pm$std accuracy; best per row in \textbf{bold}. IGD, MGD$^3$, and LGD used a pretrained DiT as the diffusion backbone.}
\label{tab:nette-woof}
\scriptsize
\setlength{\tabcolsep}{2.5pt}   
\renewcommand{\arraystretch}{0.92} 
\resizebox{\linewidth}{!}{%
\begin{tabular}{llc|cccccccc|c}
\toprule
Dataset & Model & IPC &
Random & DM \cite{zhao2023distribution}& IDC-1 \cite{kim2022dataset} & DiT \cite{peebles2023scalable} & Minimax \cite{gu2024efficient} &
IGD \cite{chen2025igd} & MGD$^3$ \cite{chan-santiago2025mgd3}& \textbf{LGD (Ours)}  & Full \\
\midrule
\multirow{6}{*}{Nette}
& \multirow{2}{*}{ConvNet-6}
& 50  & 71.8$\pm$1.2 & 70.3$\pm$0.8 & 72.4$\pm$0.7 & 74.1$\pm$0.6 & 76.9$\pm$0.9 & 80.9$\pm$0.9 & 80.9$\pm$2.3  & \textbf{82.6$\pm$0.7}  & \multirow{2}{*}{94.3$\pm$0.5} \\
& & 100 & 79.9$\pm$0.8 & 78.5$\pm$0.8 & 80.6$\pm$1.1 & 78.2$\pm$0.3 & 81.1$\pm$0.3 & 84.5$\pm$0.7 & 86.5$\pm$0.9 &\textbf{87.2$\pm$0.7}  & \\
\cmidrule(lr){2-11}
& \multirow{2}{*}{ResNetAP-10}
& 50  & 77.3$\pm$1.0 & 76.7$\pm$1.0 & 77.4$\pm$0.7 & 76.9$\pm$0.5 & 78.2$\pm$0.7 & 81.0$\pm$1.2 &81.2$\pm$1.0 &\textbf{84.3$\pm$0.5}  & \multirow{2}{*}{94.6$\pm$0.5} \\
& & 100 & 81.1$\pm$0.6 & 80.9$\pm$0.7 & 81.5$\pm$1.2 & 80.1$\pm$1.1 & 81.3$\pm$0.9 & 85.2$\pm$0.5 & 85.5$\pm$1.0  & \textbf{87.2$\pm$0.9}  & \\
\cmidrule(lr){2-11}
& \multirow{2}{*}{ResNet-18}
& 50  & 75.8$\pm$1.1 & 75.0$\pm$1.0 & 77.8$\pm$0.7 & 75.2$\pm$0.9 & 78.1$\pm$0.6 & 81.0$\pm$0.7 & 81.5$\pm$3.4  & \textbf{85.0$\pm$0.9}  & \multirow{2}{*}{95.3$\pm$0.6} \\
& & 100 & 82.0$\pm$0.4 & 81.5$\pm$0.4 & 81.7$\pm$0.8 & 77.8$\pm$0.6 & 81.3$\pm$0.7 & 84.4$\pm$0.8 & 85.6$\pm$0.2  &\textbf{86.9$\pm$0.6} & \\
\midrule
\multirow{6}{*}{Woof}
& \multirow{2}{*}{ConvNet-6}
& 50  & 41.9$\pm$1.4 & 43.8$\pm$1.1 & 42.6$\pm$0.9 & 48.5$\pm$1.3 & 50.7$\pm$1.8 & \textbf{54.2$\pm$0.7} & 53.4$\pm$0.4  & 53.9$\pm$2.2 & \multirow{2}{*}{85.9$\pm$0.4} \\
& & 100 & 52.3$\pm$1.5 & 50.1$\pm$0.9 & 51.0$\pm$1.1 & 54.2$\pm$1.5 & 57.1$\pm$1.9 & 61.1$\pm$1.0 & 59.2$\pm$0.9& \textbf{61.9$\pm$2.0}  & \\
\cmidrule(lr){2-11}
& \multirow{2}{*}{ResNetAP-10}
& 50  & 50.1$\pm$1.6 & 47.8$\pm$1.2 & 49.3$\pm$0.9 & 55.8$\pm$1.1 & 59.8$\pm$0.8 & \textbf{62.7$\pm$1.2} & 59.4$\pm$0.2 & 62.5$\pm$0.7  & \multirow{2}{*}{87.2$\pm$0.6} \\
& & 100 & 59.2$\pm$0.9 & 59.8$\pm$1.3 & 56.4$\pm$0.5 & 62.5$\pm$0.9 & 66.8$\pm$1.2 & 69.7$\pm$0.9 & 66.1$\pm$0.8 & \textbf{71.1$\pm$0.8}  & \\
\cmidrule(lr){2-11}
& \multirow{2}{*}{ResNet-18}
& 50  & 54.0$\pm$0.8 & 53.9$\pm$0.7 & 54.5$\pm$1.0 & 57.4$\pm$0.7 & 60.5$\pm$0.5 & 62.0$\pm$1.1 & 63.9$\pm$0.3  & \textbf{65.1$\pm$0.7} & \multirow{2}{*}{89.0$\pm$0.6} \\
& & 100 & 63.6$\pm$0.5 & 64.9$\pm$0.7 & 57.7$\pm$0.8 & 62.3$\pm$0.5 & 67.4$\pm$0.7 & 70.6$\pm$1.8 & 71.3$\pm$0.5  & \textbf{72.9$\pm$0.6}  & \\
\bottomrule
\end{tabular}%
} 

\vspace{-1.0em}
\end{table*}

\textbf{Learnability Guidance.}
We modulate the diffusion model's predicted noise $\epsilon_\phi(x_t, t, y)$ with the gradient of the learnability score:
\begin{equation}
\tilde{\epsilon}_\phi(x_t, t, y)
= \epsilon_\phi(x_t, t, y)
+ \lambda \cdot \rho_t \cdot \nabla_{x_t}\mathcal{S}(x_t, y),
\label{eq:learnability_guidance}
\end{equation}
where $\lambda$ controls guidance strength and $\rho_t = \sqrt{1 - \bar{\alpha}_t} \frac{\|\epsilon_\phi(x_t, t, y)\|}{\|\nabla_{x_t}\mathcal{S}(x_t, y)\|}$ is a timestep-dependent scaling factor~\cite{chen2025igd} that normalizes guidance magnitude relative to the noise level.
This transforms diffusion synthesis into an \emph{active learning mechanism}: each increment $\increment_i$ maximizes training value for the current learner state, forming a curriculum of increasing difficulty with non-redundant supervision. \cref{fig:overall_method} (bottom) shows generated samples landing in the learnable gap between current model $\theta_i$ and reference model $\theta^*$.

\textbf{Guidance Schedule and Diversity.}
Following~\cite{chen2025igd, chan-santiago2025mgd3}, we apply guidance only during timesteps $t \in [10, 45]$ (out of 50 steps), as full-trajectory guidance degrades performance. Within this window, we apply deviation guidance~\cite{chen2025igd}: for each sample $x_t$, we subtract $\gamma \nabla_{x_t} \mathcal{G}_D(x_t)$ from the predicted noise. Let $\mathcal{M}^c$ denote the memory buffer of previously generated samples for class $c$. The guidance objective $\mathcal{G}_D(x) = \frac{x \cdot \tilde{x}^*}{\|x\| \|\tilde{x}^*\|}$ measures cosine similarity to the nearest sample $\tilde{x}^* \in \mathcal{M}^c$, and $\gamma$ controls repulsion strength. This enhances intra-class diversity by steering generation away from existing samples.

\subsection{Learnability Sample Selection}
\label{sec:learnability_selection}
While learnability guidance steers diffusion trajectories toward informative regions during synthesis, the stochastic nature of sampling can still produce a mix of highly and weakly learnable examples.
To refine the distilled dataset, we introduce a per-sample learnability selection step applied after generation: for each sample to be added to the dataset, we generate $\kappa$ candidate versions via learnability-guided diffusion sampling, score each using \cref{eq:learnability_score}, and retain only the highest-scoring candidate. This selected sample is then added to the memory buffer $\mathcal{M}^c$, and the process repeats sequentially for subsequent samples, ensuring each new addition is selected in the context of the already-constructed dataset.

\textbf{Candidate Generation and Selection.}
At each incremental stage $i$, we fill $N_i$ sample positions per class by generating $\kappa$ candidates $\{(x_{j,k}^c, y_{j,k}^c)\}_{k=1}^{\kappa}$ for each position $j \in {1, \dots, N_i}$ via learnability-guided diffusion sampling. Each candidate is scored using \cref{eq:learnability_score}, and we retain the top-scoring sample:
\begin{equation}
(x_j^c, y_j^c) = \operatorname{Top}1\big({(x{j,k}^c, y{j,k}^c, \mathcal{S}(x_{j,k}^c, y_{j,k}^c))}_{k=1}^{\kappa}\big)
\end{equation}
The selected samples are appended to the memory buffer $\mathcal{M}^c$ and repeated for all $N_i$ positions to form $\increment_i$, encouraging diversity while ensuring high-learnability increments aligned with \cref{eq:increment_objective}.

\section{Experiments}
\label{sec:exp}

\captionsetup[figure]{aboveskip=2pt,belowskip=2pt}
\captionsetup[subfigure]{skip=2pt}
\begin{figure*}[tbh]
  \centering
  \begin{subfigure}[t]{0.333\linewidth}\centering
    \includegraphics[width=\linewidth, trim=6pt 5pt 0pt 7pt,clip]{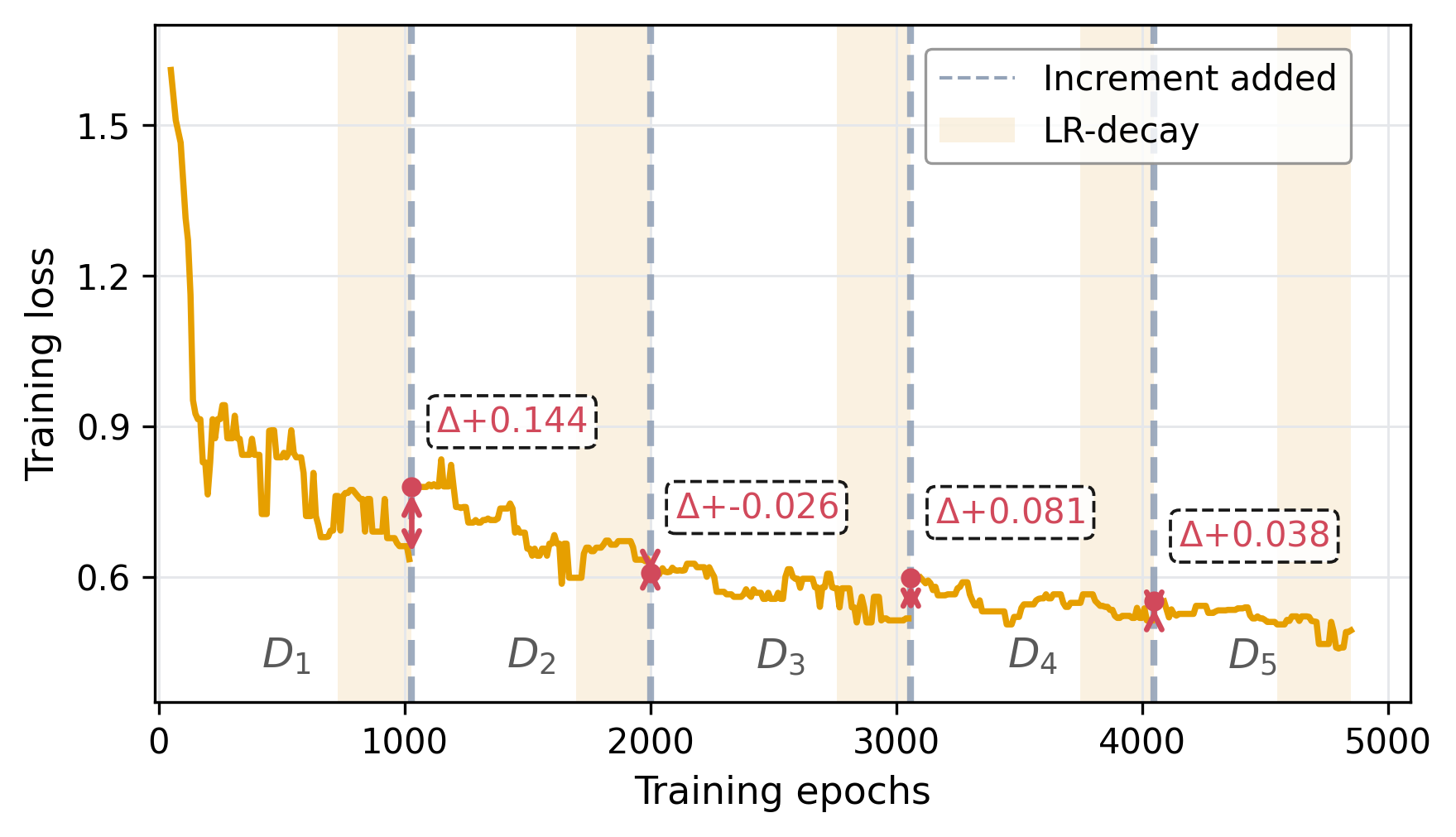}
    \caption{DiT Incremental Training}
    \label{fig:dit-loss}
  \end{subfigure}
  \begin{subfigure}[t]{0.333\linewidth}\centering
    \includegraphics[width=\linewidth, trim=6pt 5pt 0pt 7pt,clip]{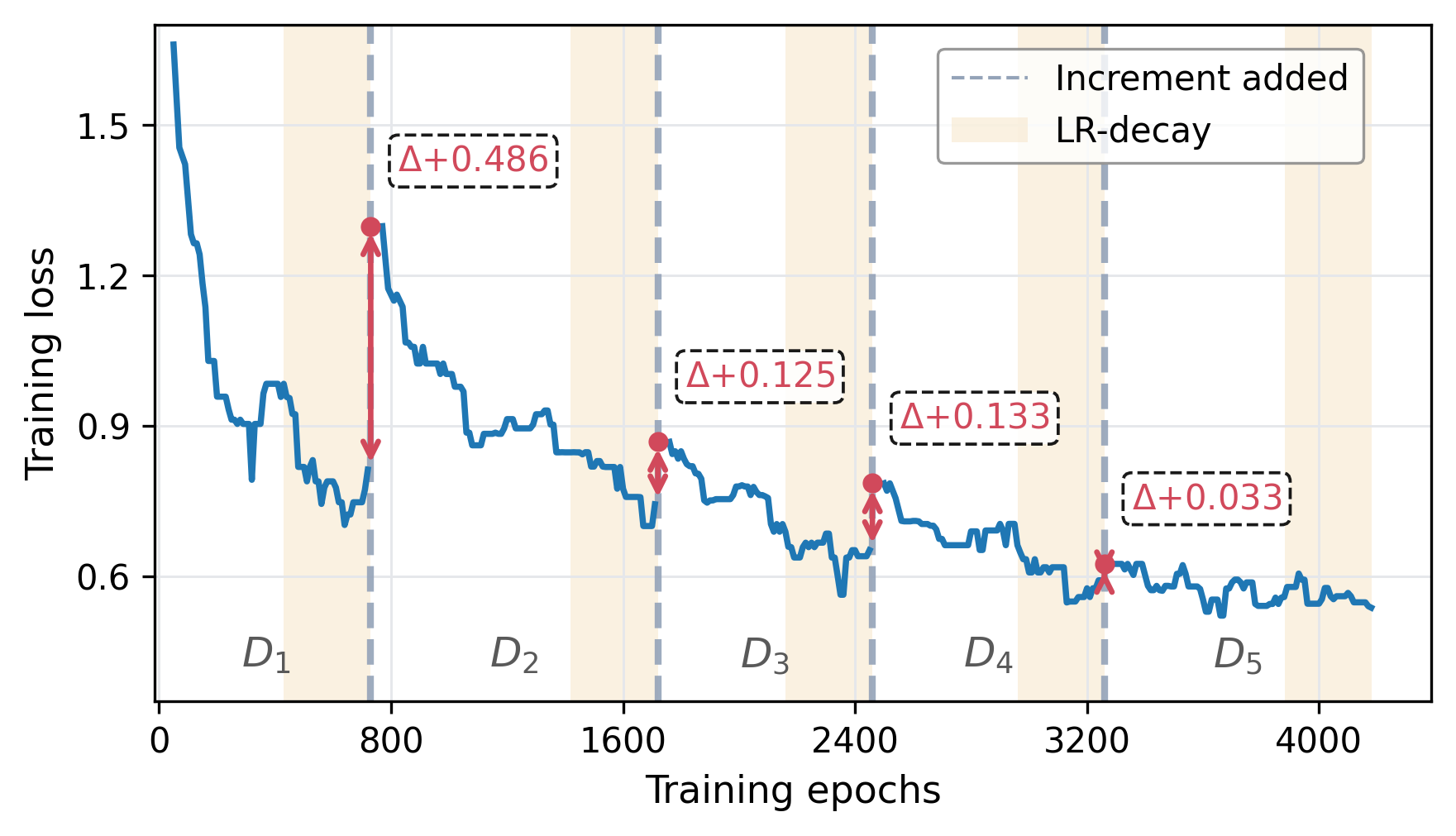}
    \caption{Ours Incremental Training}
    \label{fig:ours-loss}
  \end{subfigure}
  \begin{subfigure}[t]{0.333\linewidth}\centering
    \includegraphics[width=\linewidth, trim=6pt 5pt 0pt 7pt,clip]{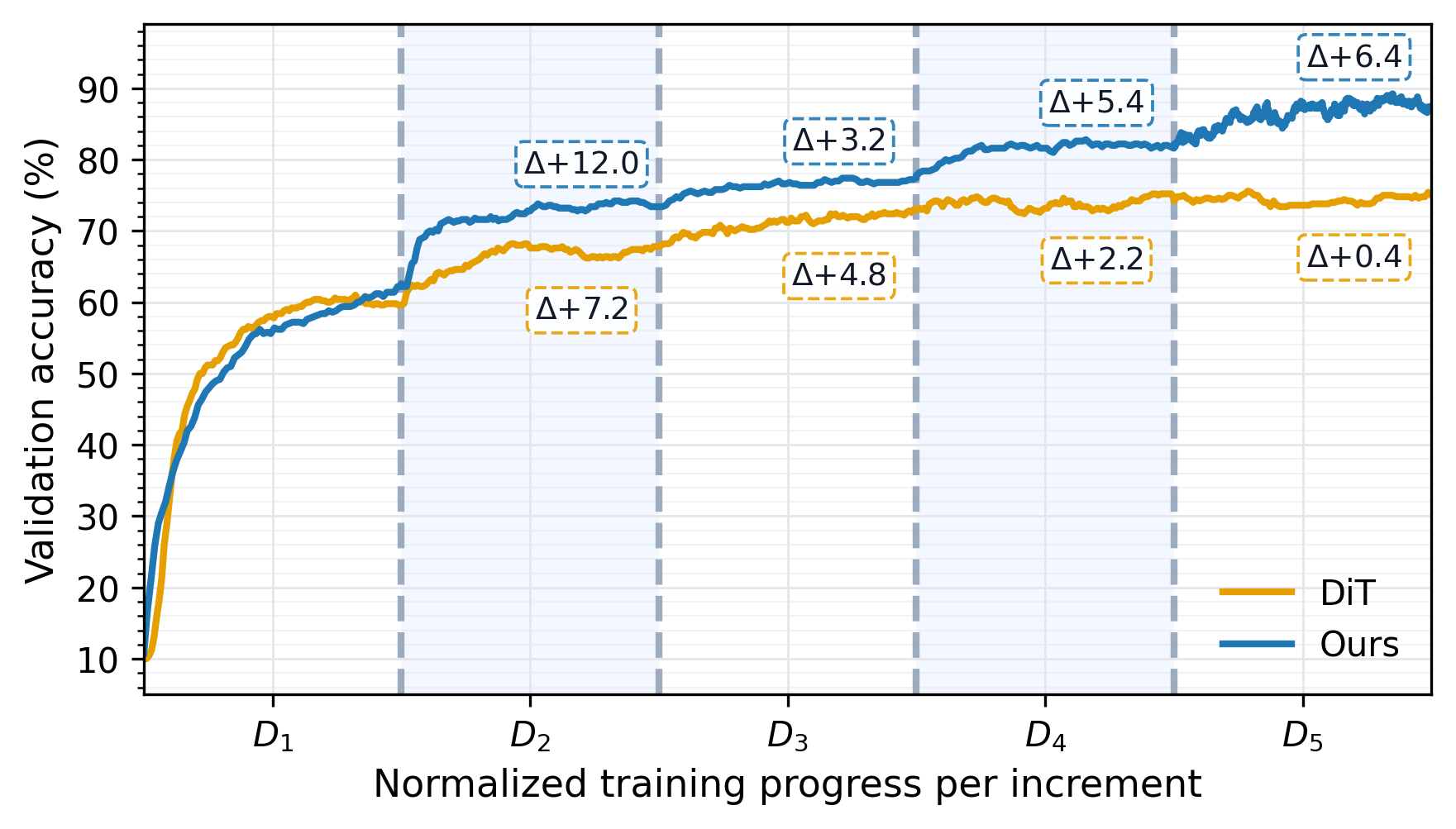}
    \caption{Validation Accuracy Progression Comparison}
    \label{fig:val}
  \end{subfigure}

  \vspace{2pt} 
  \caption{\textbf{Incremental training dynamics of DiT and our method.}
  (a-b) show the training loss across successive data increments ($D_1 \rightarrow D_5$),
  where each increment adds new samples followed by a 300-epoch learning-rate decay (light beige).
  Our method yields stronger loss spikes ($\Delta$) after each increment, suggesting the added data is harder and complementary.
  (c) compares normalized validation accuracy per increment between DiT and our method,
  highlighting consistent accuracy gains and faster convergence for ours.}
  \label{fig:progressive}
  \vspace{-1em}
\end{figure*}

\textbf{Datasets.} We evaluate on three $256 \times 256$ datasets: ImageNette~\cite{Howard_Imagenette_2019}, ImageWoof~\cite{Howard_Imagewoof_2019}, and ImageNet-1K \cite{deng2009imagenet}, in various IPC from 20 to 100.

\textbf{Protocols.} For ImageNette/ImageWoof, we use the hard-label protocol~\cite{gu2024efficient}: train target networks (ConvNet-6, ResNet-AP-10, ResNet-18) from scratch on synthetic data with ground-truth labels. For ImageNet-1K, we use the soft-label protocol~\cite{sun2024rded, gu2024efficient}: each distilled image $x_i$ is divided into $M$ regions, and each region $x_{i,m}$ is paired with a soft label $y_{i,m} = \ell(\phi_{\theta_\mathcal{T}}(x_{i,m}))$ generated by a ResNet-18 teacher. All experiments use standard augmentation (random crop, CutMix) and report mean±std of top-1 accuracy over multiple runs following \cite{chan-santiago2025mgd3, chen2025igd, gu2024efficient}.

\textbf{Hyperparameters.} We set the learnability guidance strength $\lambda = 15$, reference model weight $\omega = 0.5$, and over-generation factor $\kappa = 3$ for selection. Guidance is applied during timesteps $t \in [10, 45]$ (out of 50 diffusion steps). For deviation guidance, we use $\gamma = 50$. The initial seed dataset uses 10 IPC per class. Ablation studies and further details are in the supplementary material.

\textbf{Evaluation Settings.} We evaluate under two complementary protocols:
\textit{Static Evaluation} trains models from scratch on the complete distilled dataset (standard evaluation). 
\textit{Incremental Distillation} constructs datasets incrementally (10 IPC per stage), training models on cumulative data after each increment to analyze curriculum effects. 
For our method, incremental construction yields the final dataset; baselines distill each increment independently.

\subsection{Main Results: Static Evaluation}
\label{sec:static}
We compare our incrementally constructed datasets with current SOTA methods under standard static training protocols. Models train from scratch on the complete distilled dataset across three benchmarks, enabling direct comparison with prior work.

\begin{table}
\centering
\setlength{\tabcolsep}{3pt} 
\caption{\textbf{ImageNet-1K:} Performance comparison over ResNet-18 with state-of-the-art
dataset distillation methods.}
\vspace{-.5em}
\label{tab:imagenet1k_resnet18}
\resizebox{\columnwidth}{!}{%
\begin{tabular}{cccccccc}
\toprule
 SRe$^{2}$L \cite{yin2023sre2l} & G-VBSM \cite{li2024generative} & RDED \cite{sun2024rded} & DiT \cite{peebles2023scalable} & Minimax \cite{gu2024efficient}  & DiT-IGD \cite{chen2025igd} &  MGD$^3$\cite{chan-santiago2025mgd3}& DiT+LGD (Ours) \\
\midrule
 46.8$\pm$0.2 & 51.8$\pm$0.4 & 56.5$\pm$0.1 & 52.9$\pm$0.6 & 58.6$\pm$0.3 & 59.8$\pm$0.3 & 60.2$\pm$0.1 & 60.1$\pm$0.1  \\ 
\bottomrule
\end{tabular}
}
\vspace{-0.5em}
\end{table}
\begin{table}
\centering
\setlength{\tabcolsep}{6pt}
\caption{\textbf{Cross-Architecture Evaluation on ImageNette.}
Accuracy (\%) when models trained on surrogate architectures are evaluated on different target architectures. 
Bold indicates the best within each IPC.}
\vspace{-.5em}
\resizebox{\columnwidth}{!}{%
\begin{tabular}{l l ccc}
\toprule
 & & \multicolumn{3}{c}{\textbf{Evaluated on}} \\
\cmidrule(lr){3-5}
\textbf{IPC} & \textbf{Surrogate}  & \textbf{ConvNet-6} & \textbf{ResNetAP-10} & \textbf{ResNet-18} \\
\midrule
\multirow{3}{*}{50}
& ConvNet-6   & \textbf{83.5$\pm$0.4} & 83.5$\pm$0.8 & 83.3$\pm$1.0 \\
& ResNetAP-10 & 82.6$\pm$0.7 & \textbf{84.7$\pm$1.2} & \textbf{85.0$\pm$0.9} \\
& ResNet-18   & 81.1$\pm$1.1 & 82.3$\pm$0.3 & 83.6$\pm$1.0 \\
\midrule
\multirow{3}{*}{100}
& ConvNet-6   & 86.0$\pm$0.6 & 86.7$\pm$0.9 & 86.6$\pm$0.6 \\
& ResNetAP-10 & \textbf{87.2$\pm$0.7} & \textbf{87.7$\pm$0.6} & 86.9$\pm$0.6 \\
& ResNet-18   & 86.7$\pm$1.6 & 87.3$\pm$1.2 & \textbf{87.7$\pm$0.7} \\
\bottomrule
\end{tabular}%
}
\label{tab:cross-nette}
\vspace{-0.5em}
\end{table}
\begin{table}[t]
\centering
\footnotesize
\setlength{\tabcolsep}{3.8pt}
\renewcommand{\arraystretch}{1.15}
\caption{\textbf{Progression across data increments on ImageNette.} 
Accuracy at each IPC level as additional increments are sequentially added (IPC10 $\rightarrow$ IPC100).}
\label{tab:progressive}
\vspace{-.5em}
\resizebox{\linewidth}{!}{
\begin{tabular}{l c c c c c c}
\toprule
\textbf{Method} & \textbf{IPC10} & \textbf{IPC20} & \textbf{IPC30} & \textbf{IPC40} & \textbf{IPC50} & \textbf{IPC100} \\
\midrule
DiT &
61.0$\pm$1.5 &
68.2$\pm$1.1 &
73.0$\pm$1.7 &
75.2$\pm$1.8 &
75.6$\pm$1.1 &
78.2$\pm$0.7 \\

IGD  &
\textbf{64.5$\pm$0.9} &
71.6$\pm$1.3 &
74.8$\pm$1.7 &
78.1$\pm$1.2 &
79.9$\pm$0.3 &
84.3$\pm$0.1\\

LGD (Ours) &
64.1$\pm$1.3 &
\textbf{75.9$\pm$0.3}  &
\textbf{79.3$\pm$1.6} &
\textbf{83.1$\pm$0.8} &
\textbf{85.2$\pm$0.8} &
\textbf{89.1$\pm$0.7} \\
\bottomrule
\end{tabular}}
\vspace{-1em}
\end{table}


\begin{figure}[t]
    \centering
    \includegraphics[width=\linewidth]{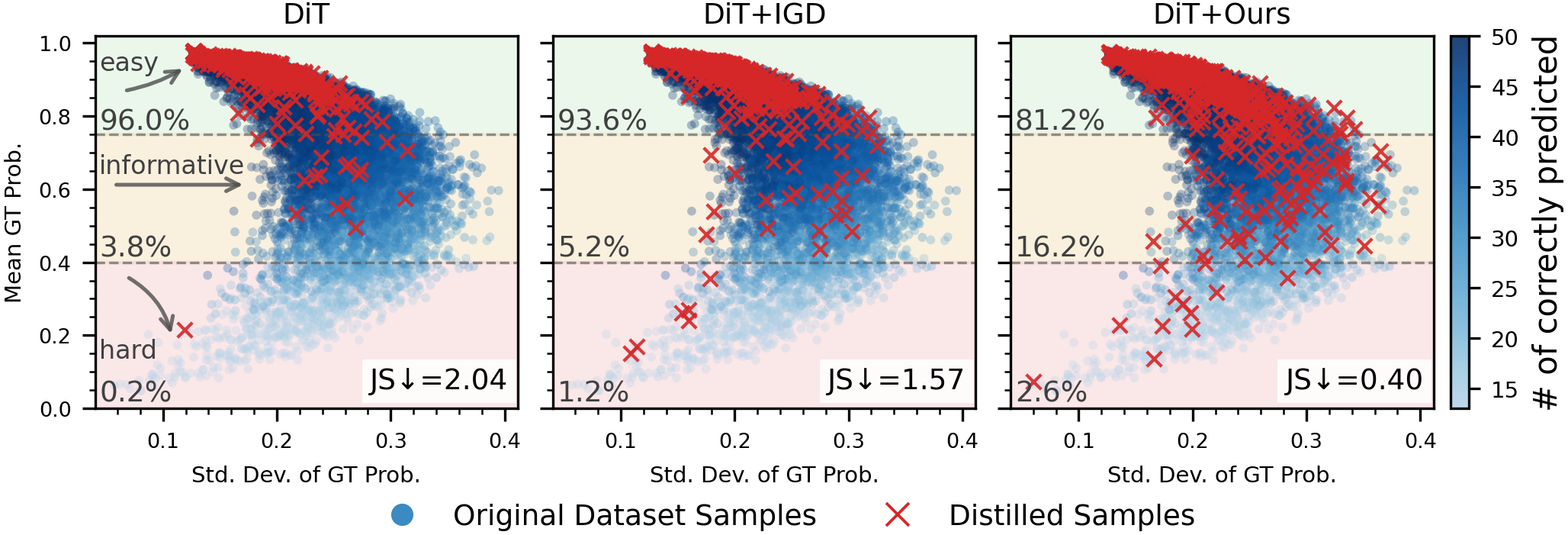}
    \caption{
\textbf{Learning-dynamics visualization of original and distilled samples.}
Each point shows a sample's mean and standard deviation of ground-truth class probability across training (50 epochs). 
Top-left points are easy (high $\mu$, low $\sigma^2$), bottom-left are hard, and mid-right indicate informative samples. 
Our method yields distilled samples that form a more informative (16.2\%) and harder dataset (2.6\%), roughly 3$\times$ and 2$\times$ over IGD, respectively, aligning more closely with the original training dynamics distribution, as shown by the lowest Jensen--Shannon divergence (JS $\downarrow$).
}
\label{fig:train-dyn}
\vspace{-1.5em}
\end{figure}

\textbf{ImageNette and ImageWoof.}
On the ImageNette dataset, our method achieves notable performance gains of $1.7\%$, $2.0\%$, $2.7\%$, and $4.0\%$ across various IPC values and architectures, surpassing DiT+IGD (see \cref{tab:nette-woof}). At 50 IPC, we reach 82.6-85.0\% across ConvNet-6, ResNet-AP-10, and ResNet-18, while at 100 IPC, we achieve 86.9-87.2\%. Similarly, on the ImageWoof dataset, our method demonstrates competitive performance at 50 IPC (53.9-65.1\%) and achieves the best results at 100 IPC (61.9-72.9\%), consistently outperforming DiT+IGD and DiT+MGD$^3$.

\textbf{ImageNet-1K.}
\cref{tab:imagenet1k_resnet18} shows comparison to SOTA in ImageNet-1K at 50 IPC with ResNet-18. Our method achieves 60.1\%, matching state-of-the-art performance (MGD³: 60.2\%) while substantially outperforming the base diffusion approach (DiT: 52.9\%, improvement of $7.2\%$) and decouple methods. This demonstrates effective scaling to complex, large-scale datasets despite incremental construction.

\begin{figure*}[t!]
    \centering
    \includegraphics[width=\linewidth]{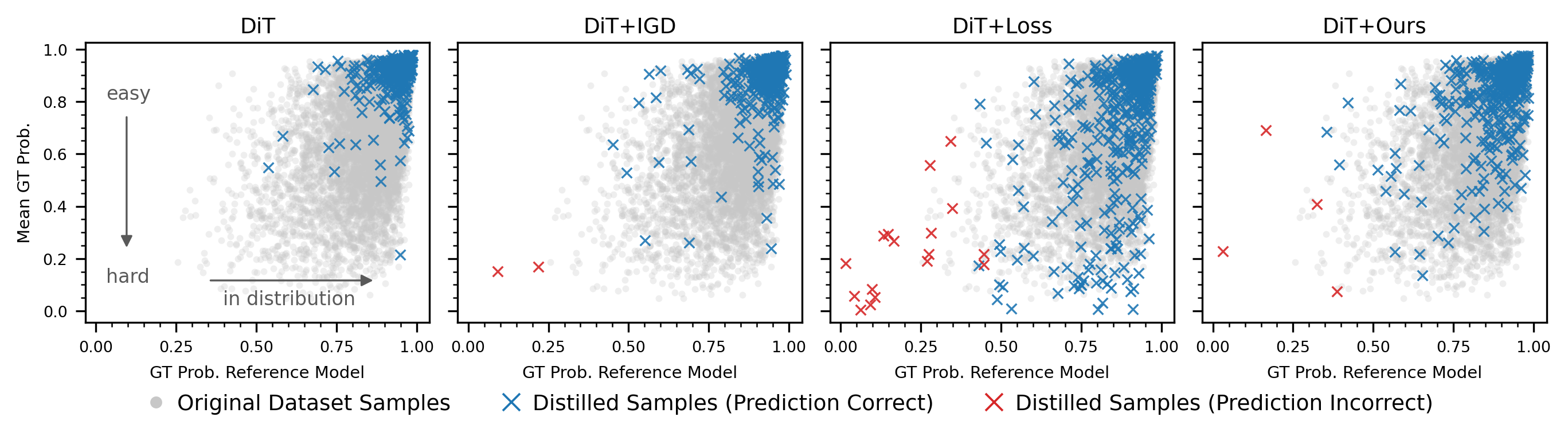}
\caption{
\textbf{In-distribution and learning-dynamics analysis of distilled datasets.}
Each point represents a sample described by its ground-truth (GT) class probability from the reference model ($x$-axis, measuring in-distribution likelihood $p(y|x)$)
and its mean GT probability across training epochs ($y$-axis, reflecting sample difficulty).
Original samples (gray) delineate the in-distribution region, where higher $y$ indicates easier examples.
\textbf{DiT} concentrates on easy, high-confidence samples;
\textbf{DiT+IGD} extends slightly toward mid and hard areas;
\textbf{DiT+Loss} covers a broader difficulty range but introduces several non-representative (low-$x$) samples;
\textbf{DiT+Ours} achieves a balanced spread—capturing informative mid and hard examples while remaining closely aligned with the in-distribution region.
}
\label{fig:diff-dist}
\vspace{-1.0em}
\end{figure*}

\textbf{Cross-Architecture Generalization.}
\cref{tab:cross-nette} shows comparison of cross-architecture transfer on ImageNette. Datasets distilled with ResNet-AP-10 achieve 85.0\% when evaluated on ResNet-18 at 50 IPC, demonstrating strong architecture-agnostic performance. At 100 IPC, cross-architecture accuracy remains high (86.0-87.7\%) with minimal degradation across all architecture pairings, confirming our method learns transferable representations rather than architecture-specific patterns.

Across all three datasets, our method achieves state-of-the-art or competitive performance (ImageNet-1K: 60.1\%, ImageNette: 82.6-87.2\%, ImageWoof: 53.9-72.9\%), with consistent improvements over uniform guidance methods. These results validate that our learnability-driven construction yields high-quality datasets. We next analyze why this approach succeeds by examining sample redundancy and training dynamics.

\subsection{Diagnosing Sample Redundancy}
\label{sec:redundancy}

To understand why learnability-driven synthesis works, we analyze sample redundancy in distilled datasets. We partition each 50 IPC dataset into $K=5$ disjoint 10 IPC increments and measure cross-increment learning: training on one increment and evaluating on another. High cross-increment accuracy indicates redundancy (overlapping information); low accuracy indicates complementary information.

\cref{fig:crossval} shows cross-validation heatmaps on ImageNette. DiT exhibits severe redundancy with 91-98\% off-diagonal accuracy (average: 94.7\%)—models trained on any increment already capture most information from others. IGD improves slightly to 81-92\% (average: 87.1\%) but substantial overlap remains. Our method achieves only 17-82\% cross-increment accuracy (average: 57.65\%), confirming that increments contain complementary information. Our method reduces redundancy by 39.1\% relative to DiT, validating that learnability-guided synthesis conditioned on model state effectively diversifies training signals.

Beyond synthesis, this demonstrates our incremental formulation as a diagnostic tool for analyzing any distillation method. By partitioning existing datasets and measuring cross-increment dynamics, researchers can systematically evaluate information distribution and identify redundancy.

\subsection{Incremental Training Dynamics}
\label{sec:progressive-training}

We now examine how reduced redundancy translates to improvement. We train models incrementally on distilled datasets, adding one 10 IPC increment at a time (5 total). For each increment, we train the model until the training loss no longer improves, then apply a learning rate decay for 300 more epochs before adding a new increment.

\textbf{Loss Spikes and Information Content.}
\cref{fig:dit-loss,fig:ours-loss} shows training loss across increments. DiT exhibits small, uniform loss spikes (avg. $\Delta = 0.06$) after each increment, consistent with redundant samples. Our method shows larger loss spikes when adding new data ($\Delta_1 = 0.486$ → $\Delta_5 = 0.033$, avg. $\Delta = 0.20$), confirming increments of increasing difficulty. Despite larger disruptions, our method converges faster to higher accuracy.

\textbf{Sustained Marginal Gains.}
\cref{fig:val} and \cref{tab:progressive} show accuracy progression. Our method achieves 85.2\% vs. 79.9\% (IGD) and 75.6\% (DiT) at 50 IPC. As shown in \cref{fig:val}, our method maintain sustained gains across stages (+12.0\%, +3.2\%, +5.4\%, +6.4\%) while DiT shows declining returns (+7.2\%, +4.8\%, +2.2\%, +0.4\%) . Our final increment (40→50 IPC) alone provides more gain than DiT's last two combined.

These dynamics validate our curriculum hypothesis: conditioning synthesis on model state automatically generates samples at the frontier of current capability—hard enough for new information, learnable enough for efficient absorption.

\begin{figure*}[tbh]
    \centering
    \includegraphics[page=1,width=\linewidth]{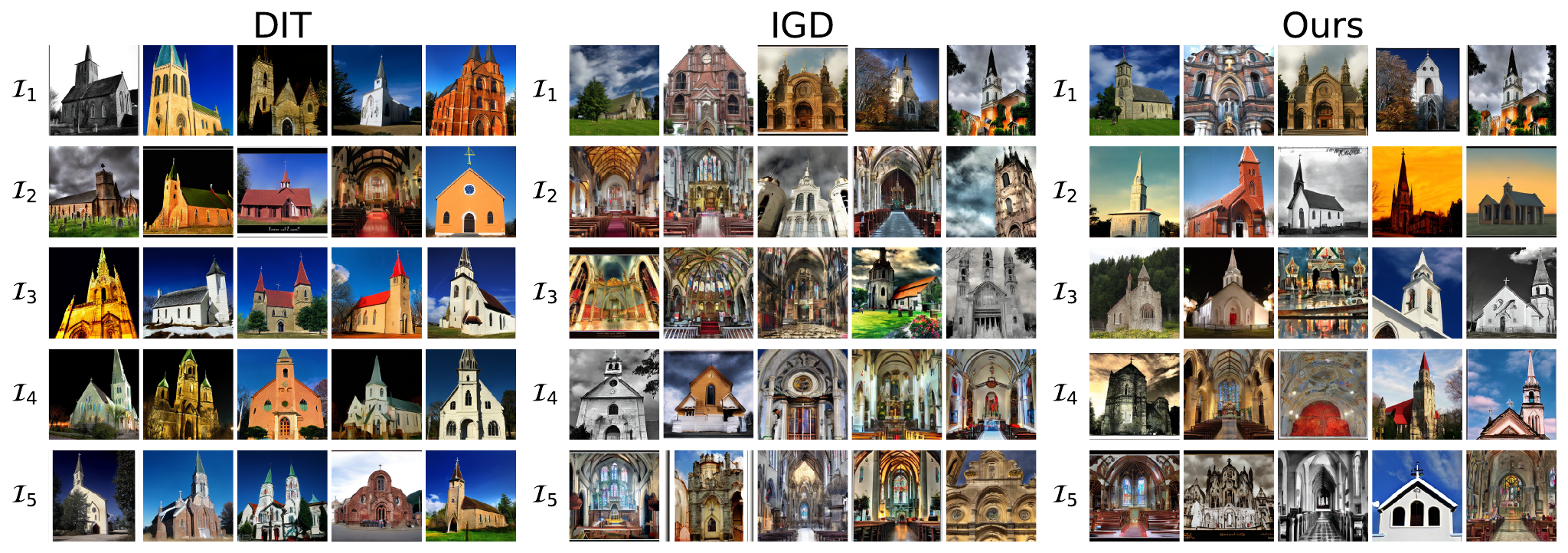}
\caption{\textbf{Visual diversity in incrementally distilled datasets.}
Samples from increments
$\increment_1  -\increment_5$ (50 IPC total) on the Church class.
\textbf{DiT} generates repetitive samples with similar architectural styles
and lighting. \textbf{IGD} improves slightly but exhibits clustering around
Gothic exteriors. \textbf{LGD (Ours)} produces diverse samples spanning multiple
architectural styles (traditional, modern, ornate), perspectives (exterior and interior views),
and lighting conditions (day, night, golden hour). This diversity reflects
our curriculum-based synthesis: early increments capture simpler structures
while later increments progressively introduce architectural complexity and
challenging interior scenes, reducing redundancy and maximizing sample utility
across the distilled dataset.}
\label{fig:church_diversity}
\end{figure*}

\subsection{Sample Quality and Difficulty Analysis}
\label{sec:sample-analysis}

We validate that our samples align with natural difficulty distributions while maintaining semantic validity.

\textbf{Difficulty Distribution.}
Following~\cite{swayamdipta-etal-2020-dataset, He_2024_CVPR}, we characterize samples by mean ($\mu$) and variance ($\sigma^2$) of GT probability across 50 training epochs. Easy samples have high $\mu$, low $\sigma^2$; hard samples have low $\mu$, low $\sigma^2$; informative samples have mid-to-high $\mu$ and high $\sigma^2$ (model improves during training).

\cref{fig:train-dyn} shows ($\mu$, $\sigma^2$) distributions. DiT concentrates in easy regions ($>$80\% with $\mu > 0.8$), IGD improves (~60\% easy) but retains clustering, while our method distributes broadly across easy, informative (high $\sigma^2$), and hard regions. Jensen-Shannon divergence quantifies alignment: our method achieves \textbf{JS = 0.40} vs. DiT (2.04) and IGD (1.57)—5× and 4× improvements, confirming better alignment with natural data difficulty characteristics.

\textbf{In-Distribution Analysis and Semantic Consistency}
\cref{fig:diff-dist} characterizes samples by reference model confidence $p(y|x)$ (x-axis, in-distribution likelihood) and mean GT probability at training (y-axis, difficulty). DiT clusters in easy regions (top-right). We also evaluate our method without the $\mathcal{L}(\theta^*, x, y)$ regularization term (denoted as DiT+Loss). While this covers a broader difficulty range, it introduces out-of-distribution samples that are misclassified by the reference model (red × markers at low x-values). Our method achieves balanced spread across difficulty levels ($x > 0.6$ for most samples), with red × appearing in genuinely hard regions (bottom-right) rather than out-of-distribution areas. This validates our learnability regularization (\cref{eq:increment_objective}): the $\mathcal{L}(\theta^*, x, y)$ term prevents out-of-distribution drift while enabling hard in-distribution samples. These analyses confirm our method spans natural difficulty ranges without semantic drift, explaining the sustained improvements in \cref{sec:progressive-training}.

\textbf{Qualitative Results.}
To demonstrate how our curriculum-based approach introduces diversity progressively, we visualize samples from each increment $\increment_1  -\increment_5$ for the Church class in \cref{fig:church_diversity}.
Compared to DiT's repetitive samples and IGD's clustering around similar structures, our method progressively introduces diversity across increments.
Early increments capture simpler, easily learned structures that establish foundational features, while later increments introduce architectural complexity and challenging scenes.
This curriculum design ensures that each increment contributes unique visual information rather than duplicating patterns from previous increments, maximizing sample utility across the distilled dataset.

\section{Conclusion}

Existing dataset distillation methods suffer from substantial redundancy, with disjoint subsets capturing 80--90\% overlapping training signals. This arises because methods optimize for visual diversity or average trajectories without considering training signal complementarity across samples.
We propose learnability-driven dataset distillation that constructs synthetic datasets incrementally, conditioning each stage on the current model state to generate complementary rather than redundant samples. Our learnability-guided diffusion automatically produces curriculum-aligned samples by balancing current-model informativeness with reference-model validity. This reduces redundancy by 39.1\%, enables specialization across training phases, and achieves state-of-the-art results: ImageNet-1K (60.1\%), ImageNette (87.2\%), and ImageWoof (72.9\%). Our framework establishes foundations for adaptive distillation methods that exploit staged learning dynamics to maximize sample efficiency.

{
    \small
    \bibliographystyle{ieeenat_fullname}
    \bibliography{main}
}

\clearpage
\setcounter{page}{1}
\setcounter{section}{0}

\maketitlesupplementary
\renewcommand{\thesection}{\Alph{section}}

This supplementary material is organized as follows: \cref{supp:sec:generalizability} presents generalizability experiments across different diffusion models. \cref{supp:sec:complementarity} provides detailed cross-increment complementarity analysis demonstrating the reduced redundancy of our method. \cref{supp:sec:seed_ablation} examines the effect of different dataset initialization strategies. \cref{supp:sec:hyperparameters} presents comprehensive hyperparameter sensitivity analysis. \cref{supp:sec:batch_size} analyzes the impact of batch size for learnability ranking. \cref{supp:sec:component_ablation} provides systematic ablation studies of each component. \cref{supp:sec:incremental_algorithm} details the incremental learning algorithm. \cref{supp:sec:implementation} provides complete implementation details for reproducibility. \cref{supp:sec:qualitative} presents qualitative visualizations of generated samples across different methods.

\section{Extended Experimental Results}

This section provides comprehensive experimental analyses, including generalizability studies, redundancy analysis, ablation studies, and hyperparameter sensitivity analysis.

\subsection{Generalizability Across Base Diffusion Models}
\label{supp:sec:generalizability}

To validate that our learnability-guided framework generalizes beyond DiT~\cite{peebles2023scalable}, we apply our method to Minimax Diffusion~\cite{gu2024efficient}, a diffusion model fine-tuned for dataset distillation to encourage both diversity and representativeness. We compare three configurations: vanilla Minimax, Minimax with IGD guidance~\cite{chen2025igd}, and Minimax with our learnability-guided approach (Minimax+Ours) on ImageNette and ImageWoof at 50 and 100 IPC across three architectures.

Our method consistently improves upon both the Minimax baseline and Minimax-IGD across all architectures, datasets, and IPC budgets, as shown in \cref{supp:tab:minimax_experiments}. On both ImageNette and ImageWoof, Minimax+Ours matches or exceeds the strongest baseline in almost every setting. These consistent improvements demonstrate that our learnability-guided synthesis framework is architecture-agnostic and can be seamlessly integrated with different generative priors to enhance dataset distillation quality.

\begin{table}[tbh]
\centering
\caption{Comparison across distilled IPC budgets with Minimax Diffusion \cite{gu2024efficient} on Nette and Woof  Mean$\pm$std accuracy; best per row is shown in \textbf{bold}. Full represents training with the entire dataset.}
\label{supp:tab:minimax_experiments}
\scriptsize
\setlength{\tabcolsep}{2.5pt}   
\renewcommand{\arraystretch}{0.92} 
\resizebox{\linewidth}{!}{%
\begin{tabular}{llc|ccc|c}
\toprule
Dataset & Model & IPC & Minimax\cite{gu2024efficient} & Minimax-IGD\cite{chen2025igd} & Minimax+Ours  & Full \\
\midrule
\multirow{6}{*}{Nette}
& \multirow{2}{*}{ConvNet-6}
& 50  & 76.9$\pm$0.9 & \textbf{80.6$\pm$0.8} & 77.8$\pm$1.0 & \multirow{2}{*}{94.3$\pm$0.5} \\
& & 100 &  81.1$\pm$0.3  & \textbf{85.1$\pm$0.5} & \textbf{85.1$\pm$0.7}  & \\
\cmidrule(lr){2-7}
& \multirow{2}{*}{ResNetAP-10}
& 50  & 78.2$\pm$0.7 & 81.5$\pm$0.3 & \textbf{82.2$\pm$0.4} & \multirow{2}{*}{94.6$\pm$0.5} \\
& & 100 &  81.3$\pm$0.9 & 85.6$\pm$0.3 &  \textbf{86.4$\pm$0.4} & \\
\cmidrule(lr){2-7}
& \multirow{2}{*}{ResNet-18}
& 50  &  78.1$\pm$0.6  & 82.2$\pm$0.4  & \textbf{82.5$\pm$0.4} & \multirow{2}{*}{95.3$\pm$0.6} \\
& & 100 &  81.3$\pm$0.7  & 85.3$\pm$1.0 & \textbf{87.2$\pm$0.8}  & \\
\midrule
\multirow{6}{*}{Woof}
& \multirow{2}{*}{ConvNet-6}
& 50  & 50.7$\pm$1.8 & 54.6$\pm$1.3 & \textbf{56.3$\pm$1.4}  & \multirow{2}{*}{85.9$\pm$0.4} \\
& & 100 &  57.1$\pm$1.9 &  61.3$\pm$0.9 & \textbf{62.6$\pm$1.0}  & \\
\cmidrule(lr){2-7}
& \multirow{2}{*}{ResNetAP-10}
& 50  & 59.8$\pm$0.8 &  62.5$\pm$1.4 &  \textbf{63.7$\pm$1.1} & \multirow{2}{*}{87.2$\pm$0.6} \\
& & 100 &  66.8$\pm$1.2 & 68.3$\pm$0.6  & \textbf{70.0$\pm$0.5} & \\
\cmidrule(lr){2-7}
& \multirow{2}{*}{ResNet-18}
& 50  & 60.5$\pm$0.5 & 63.4$\pm$0.6 & \textbf{66.2$\pm$0.6} & \multirow{2}{*}{89.0$\pm$0.6} \\
& & 100 &  67.4$\pm$0.7 & 70.5$\pm$0.8 & \textbf{72.1$\pm$0.4}  & \\
\bottomrule
\end{tabular}%
} 

\end{table}

\subsection{Cross-Increment Complementarity Analysis}

\label{supp:sec:complementarity}

To quantify sample redundancy in distilled datasets, we partition each 50 IPC dataset into $K=5$ disjoint 10 IPC increments and measure cross-increment learning dynamics. Specifically, we train a ResNet-AP-10 model on the first increment $\mathcal{I}_1$ (10 IPC) until convergence, then evaluate classification errors on subsequent increments $\mathcal{I}_{t+1}$ for $t \in \{1, 2, 3, 4\}$, without any additional training. High error counts indicate complementary information (the model has not learned the new increment's signals), while low errors indicate redundancy (overlapping information has already been captured).

\begin{figure}[t]
    \centering
    \includegraphics[width=\linewidth]{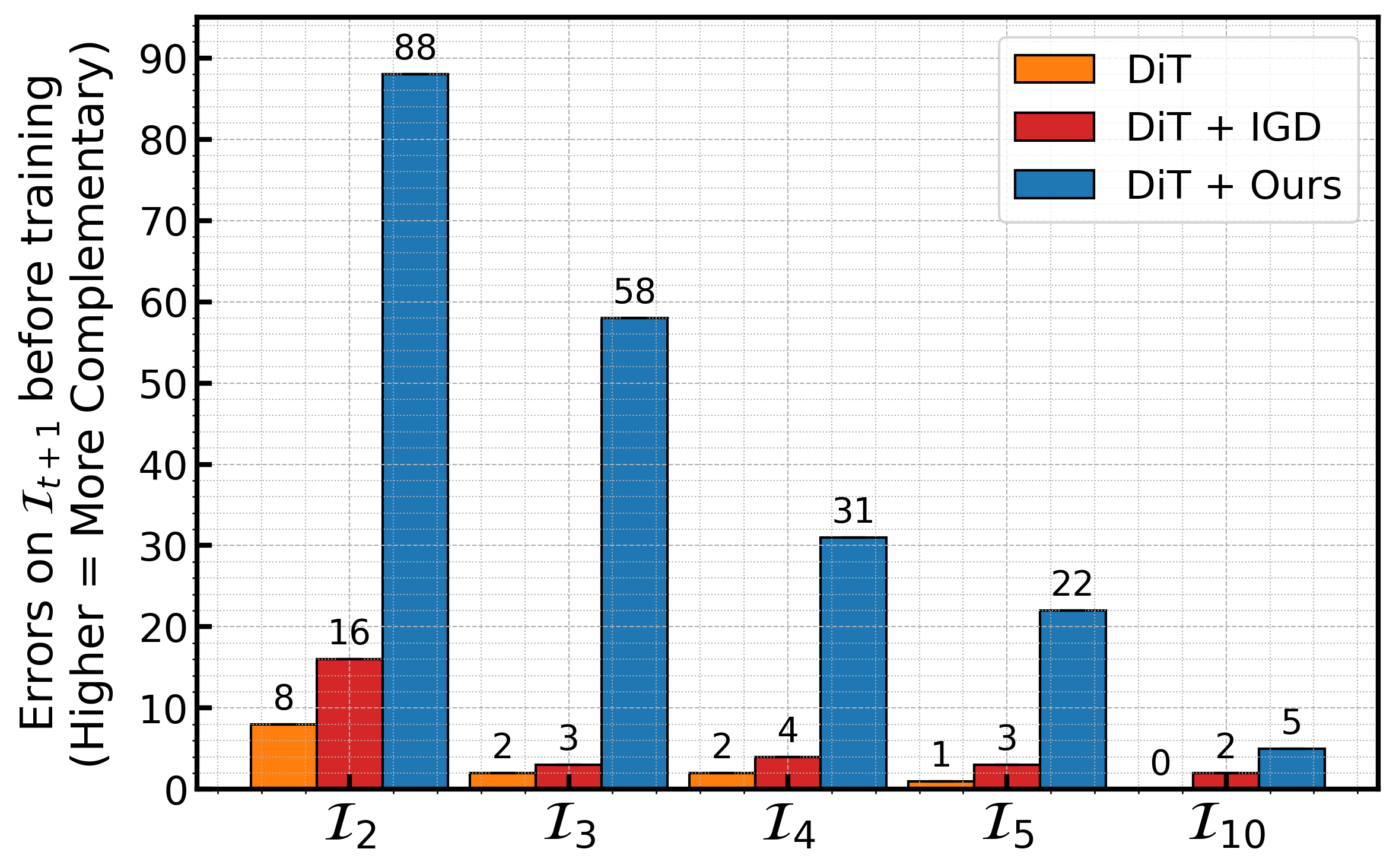}
    \caption{\textbf{Cross-increment redundancy analysis.} Errors on increment 
$\mathcal{I}_{t+1}$ using a model trained only on $\mathcal{I}_t$. DiT and IGD show $<15$ errors (high redundancy); our method shows 22-88 errors, confirming reduced redundancy.}
\label{supp:fig:cross_increment_redundancy}
\end{figure}

\cref{supp:fig:cross_increment_redundancy} shows results on ImageNette. DiT exhibits severe redundancy with only 0--8 errors across increments (92--100\% accuracy), indicating that a model trained on $\mathcal{I}_1$ already captures most of the training signal from remaining increments. IGD improves slightly with 2--16 errors (84--98\% accuracy) but substantial overlap remains. In contrast, our method maintains 22--88 errors (12--78\% accuracy), confirming that each increment introduces substantial new learning signals. Notably, our method shows a decreasing trend in errors across increments ($\mathcal{I}_2$: 88 errors, $\mathcal{I}_3$: 58 errors, $\mathcal{I}_4$: 31 errors, $\mathcal{I}_5$: 22 errors), suggesting that finding hard samples becomes progressively more challenging as the model improves; a natural consequence of curriculum learning where easier patterns are learned first. Even at $\mathcal{I}_5$ (the 40--50 IPC increment), our method produces 22 errors versus 0 for DiT, demonstrating sustained complementarity across all stages. This 5--10$\times$ higher error rate validates that our learnability-guided synthesis conditions each stage on model state to generate non-redundant, complementary samples rather than redundant copies.

\subsection{Effect of Initial Data Seed on Our Method}
\label{supp:sec:seed_ablation}

Our method begins with a seed dataset $\mathcal{D}_1$ of size 10 IPC. To understand how this initialization affects the final distilled dataset, we compare three seed sources: IGD~\cite{chen2025igd}, DiT~\cite{peebles2023scalable}, and randomly selected real images. We evaluate accuracy as the distilled dataset expands from 10 to 50 IPC on ImageNette using ResNet-AP-10, reporting results under the incremental training setup.

\begin{figure}[t]
    \centering
    \includegraphics[width=\linewidth]{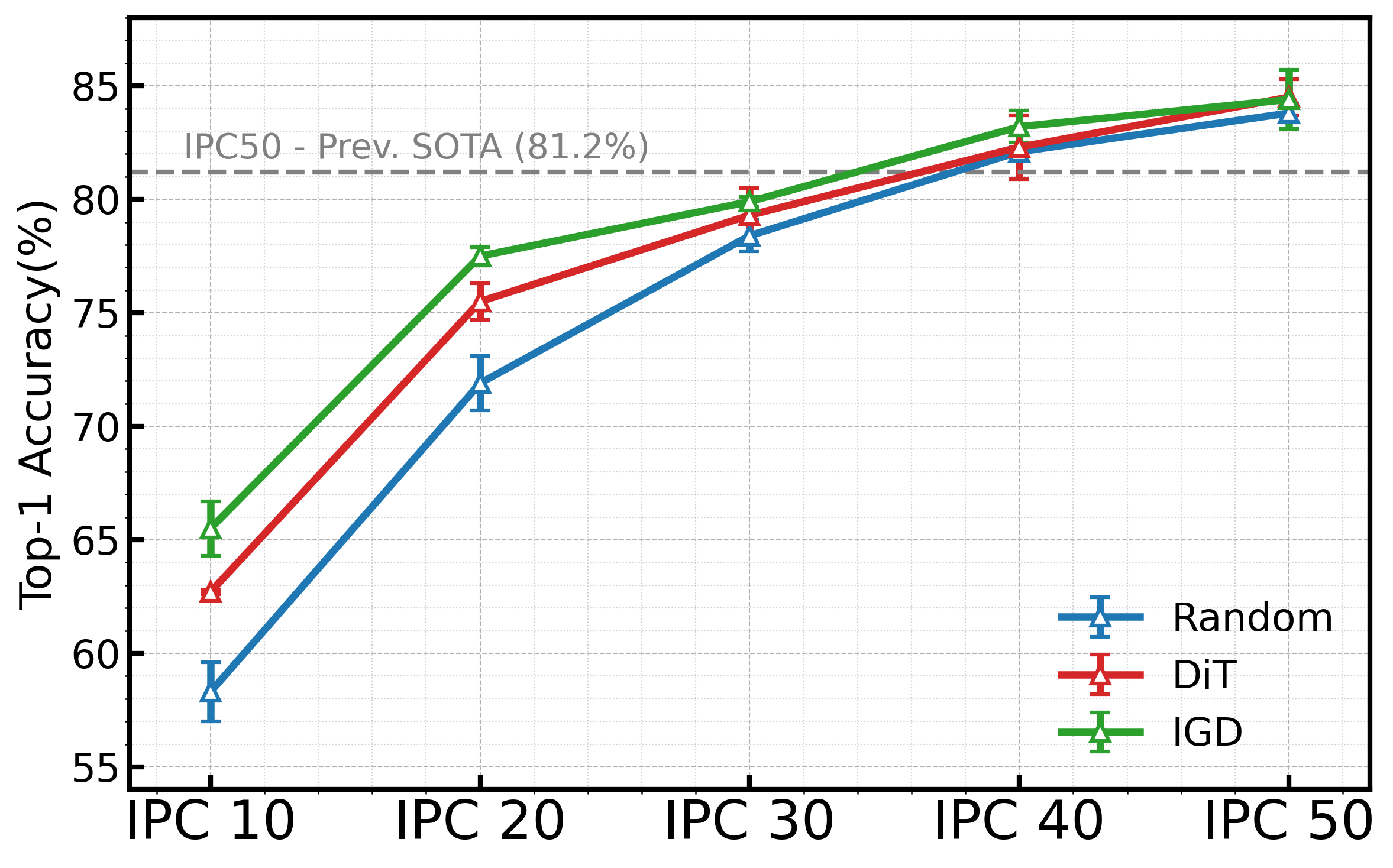}
    \caption{\textbf{Incremental accuracy across different seed initializations on our method.} Top-1 accuracy as the distilled dataset grows from 10 to 50 IPC. Three seed sources are compared: IGD~\cite{chen2025igd} (green), DiT~\cite{peebles2023scalable} (red), and random real images (blue). IGD initialization achieves the best final performance (84.4\%), but all seeds substantially outperform previous SOTA (81.2\%, dashed line), demonstrating the robustness of our learnability-guided framework to initialization choice.}
\label{supp:fig:data_init}
\end{figure}

\cref{supp:fig:data_init} shows that IGD provides the strongest starting point, improving final accuracy by 0.6\% over random initialization (84.4 vs.\ 83.8). This advantage due to the fact that   IGD's influence-guided generation produces samples aligned with informative low-IPC training gradients (10–20 IPC). DiT achieves a comparable improvement of 0.7\% over random (84.5 vs.\ 83.8). Importantly, our learnability-guided framework amplifies all initializations: even with random seeds, we reach 83.8\%. For comparison purposes, MGD$^3$ achieves 81.2\% when distilling 50 IPC under its static evaluation protocol.

While stronger seeds yield higher absolute performance, the key advantage of our method—progressive, complementary synthesis—remains effective regardless of initialization. We default to IGD, though other seeds may be preferable under tighter computational budgets. The consistent relative gains highlight that our incremental, learnability-driven formulation provides benefits largely independent of seed quality.

\subsection{Hyperparameter Analysis}
\label{supp:sec:hyperparameters}
\begin{figure*}[t]
    \centering
    \begin{subfigure}{0.33\linewidth}
        \centering
        \includegraphics[width=\linewidth]{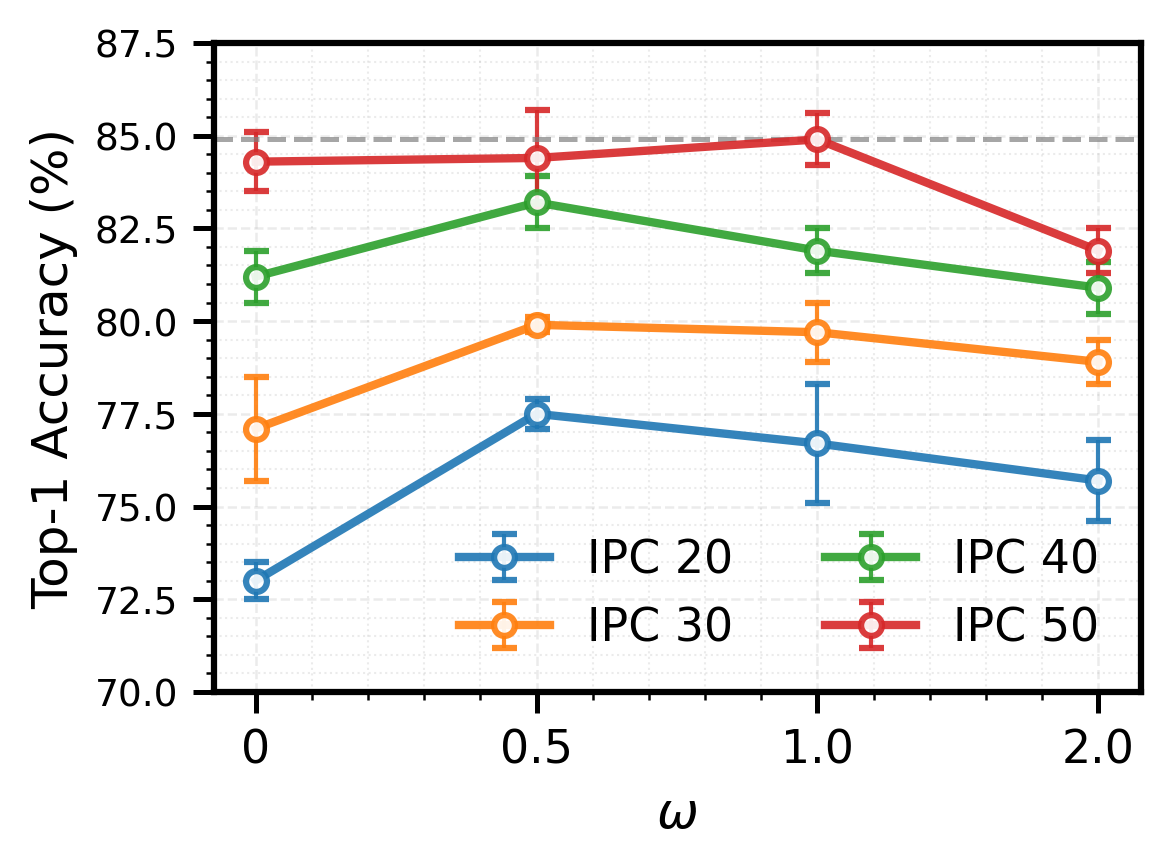}
        \caption{Effect of reference weight $\omega$.}
        \label{supp:fig:omega_ablation}
    \end{subfigure}
    \hfill
    \begin{subfigure}{0.33\linewidth}
        \centering
        \includegraphics[width=\linewidth]{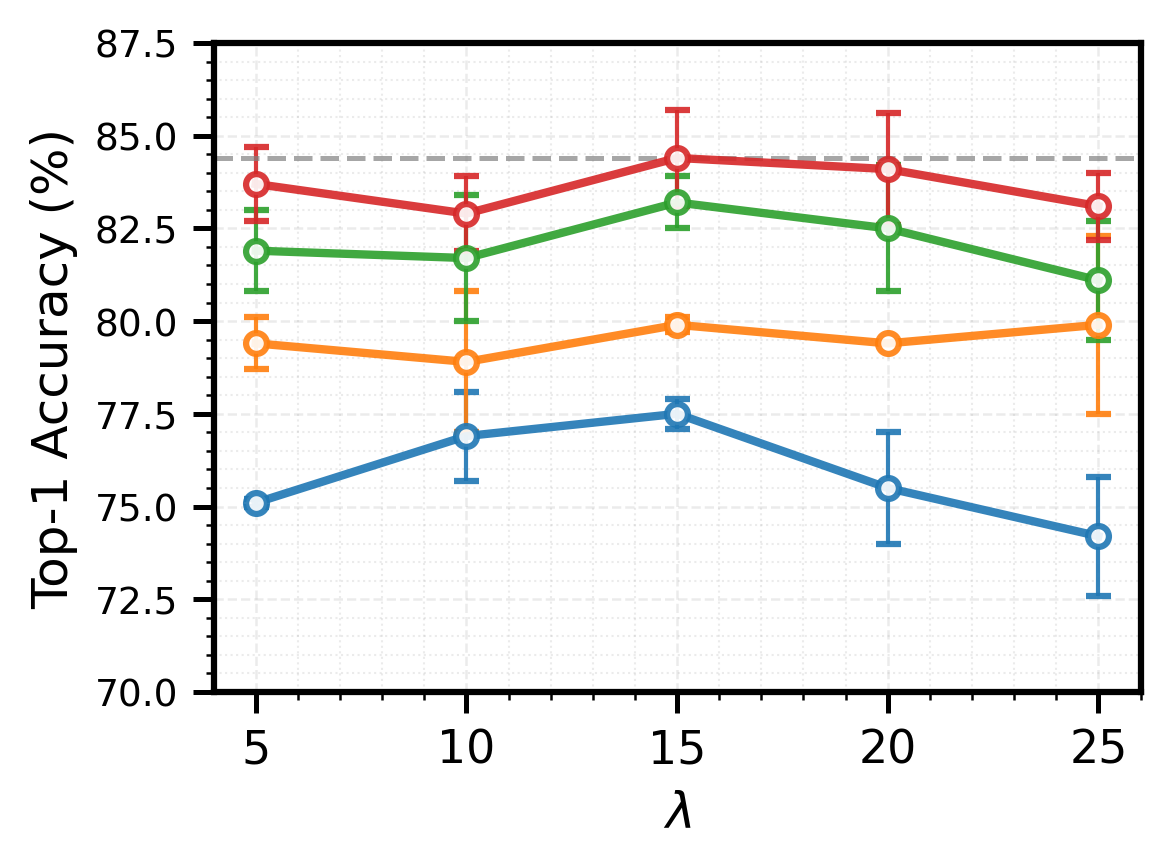}
        \caption{Effect of guidance strength $\lambda$.}
        \label{supp:fig:lambda_ablation}
    \end{subfigure}
    \hfill
    \begin{subfigure}{0.33\linewidth}
        \centering
        \includegraphics[width=\linewidth]{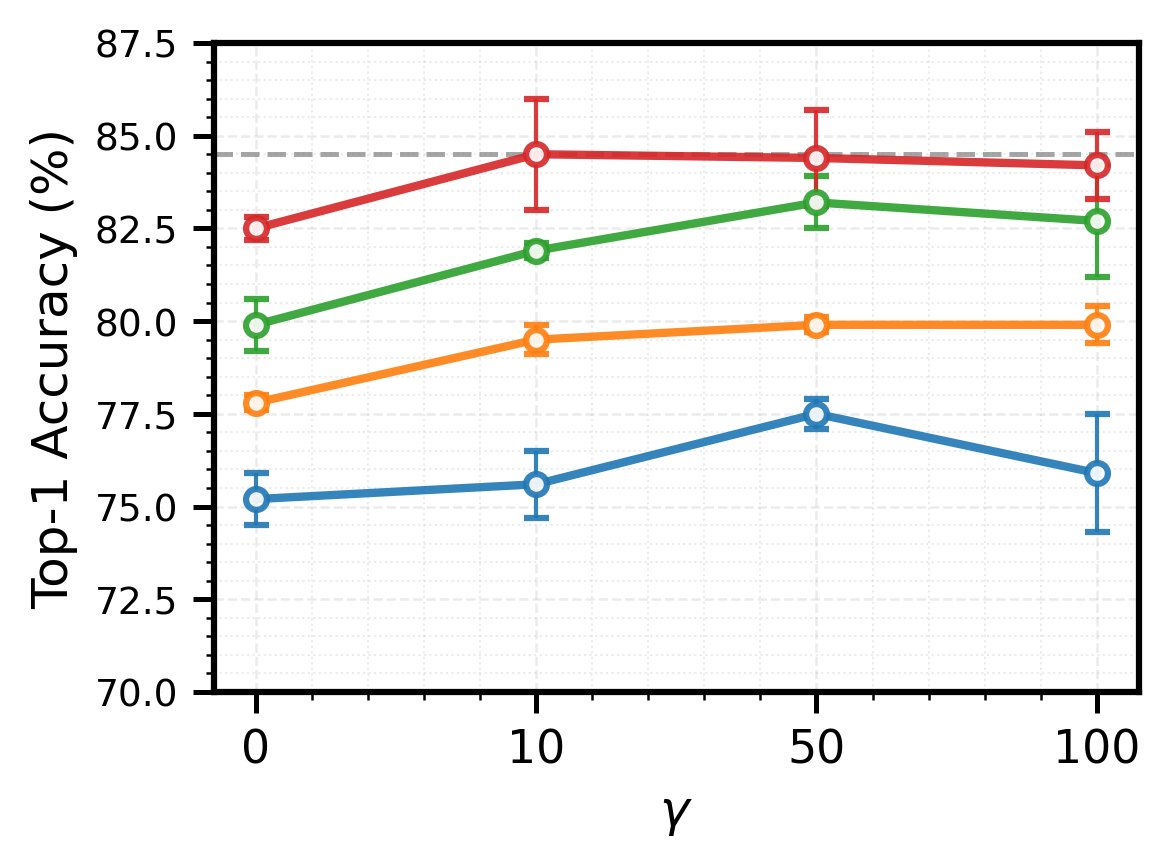}
        \caption{Effect of deviation strength $\gamma$.}
        \label{supp:fig:gamma_ablation}
    \end{subfigure}

    \caption{\textbf{Hyperparameter sensitivity on ImageNette.}
    Top-1 accuracy (\%) as we vary (a) reference weight $\omega$,
    (b) guidance strength $\lambda$, and (c) deviation strength $\gamma$
    across different IPC settings.}
    \label{fig:hyperparam_sweeps}
\end{figure*}

Our method uses three key hyperparameters: reference model weight $\omega$ (\cref{eq:learnability_score} in main paper), guidance strength $\lambda$ (\cref{eq:learnability_guidance} in main paper), and deviation guidance strength $\gamma$ (\cref{sec:learnability_guidance} in main paper). We systematically evaluate each parameter's impact on distilled dataset quality across multiple IPC budgets.

\subsubsection{Reference Model Weight  \texorpdfstring{$\omega$}{omega}}
To assess how the reference-model weight $\omega$ influences performance, we evaluate $\omega \in {0.0, 0.5, 1.0, 2.0}$ across IPC levels 20, 30, 40, and 50. As shown in \cref{supp:fig:omega_ablation}, the optimal setting is $\omega = 1.0$, achieving 84.9\% at 50 IPC. When $\omega = 0.0$ (i.e., no reference-model guidance), performance drops sharply at 20 IPC, suggesting that the resulting samples are too difficult for low-IPC training. In contrast, $\omega = 2.0$ overemphasizes the reference model and produces overly easy samples, reducing performance at higher IPC levels such as 50. Overall, the method is stable within the range $\omega \in [0.5, 1.0]$, where accuracy remains within 1\% of the optimal value.

\subsubsection{Guidance Strength \texorpdfstring{$\lambda$}{lambda}}

We evaluate guidance strength $\lambda \in \{5, 10, 15, 20, 25\}$ on ImageNette with IPC 20, 30, 40, 50, and 100. \cref{supp:fig:lambda_ablation} shows that $\lambda = 15$ achieves the best performance at 84.4\% for 50 IPC. When guidance is too weak ($\lambda = 5$), performance drops to 83.7\%. Conversely, when guidance is too strong ($\lambda = 25$), performance declines to 83.1\%.

\subsubsection{Deviation Guidance Strength \texorpdfstring{$\gamma$}{gamma}}

To understand how deviation guidance strength affects performance, we evaluate $\gamma \in \{0, 10, 50, 100\}$ across various IPC values. Our results in \cref{supp:fig:gamma_ablation} show that $\gamma = 50$ provides the best diversity-quality trade-off, achieving 84.7\% at 50 IPC. When $\gamma = 0$ (no diversity enforcement), performance drops to 82.5\%. Extending to $\gamma = 100$ slightly reduces performance to 84.2\%.

We observe consistent trends across IPC budgets, with values at $\omega=0.5$, $\lambda=15$, and $\gamma=50$. The method demonstrates robustness within reasonable ranges: $\omega \in [0.5, 1.0]$, $\lambda \in [10, 20]$, and $\gamma \in [10, 100]$ all achieve competitive performance, enabling practitioners to use default values without extensive per-dataset tuning. 


\begin{table}[t]
\centering
\caption{\textbf{Experiment with Batch Size for Learnability Rank on ImageNette.}
Top-1 accuracy (\%) for different batch sizes and IPC settings.}
\vspace{4pt}
\resizebox{\columnwidth}{!}{%
\begin{tabular}{cccccc}
\toprule
$\mathbf{\kappa}$ & \textbf{IPC 20} & \textbf{IPC 30} & \textbf{IPC 40} & \textbf{IPC 50} & \textbf{IPC 100} \\
\midrule
1  & $77.5 \pm 0.5$ & $79.9 \pm 0.3$ & $83.2 \pm 0.8$ & $84.4 \pm 1.6$ & $88.5 \pm 1.1$ \\
3  & $77.3 \pm 1.0$ & $80.9 \pm 0.7$ & $83.3 \pm 0.6$ & $85.1 \pm 0.8$ & $88.4 \pm 1.4$ \\
10 & $\mathbf{77.8 \pm 0.4}$ & $\mathbf{81.1 \pm 1.5}$ & $\mathbf{83.9 \pm 0.8}$ & $\mathbf{85.6 \pm 0.5}$ & $\mathbf{89.1 \pm 0.6}$ \\
\bottomrule
\end{tabular}
}%
\label{supp:tab:batch_size}
\end{table}

\subsection{Effect of Batch Size for Learnability Ranking}
\label{supp:sec:batch_size}

Our learnability scoring mechanism (\cref{sec:learnability_selection} in main paper) evaluates candidate samples in batches of size $\kappa$. To assess the impact of batch size, we test $\kappa \in \{1, 3, 10\}$ across IPC budgets. 

\cref{supp:tab:batch_size} shows that larger batch sizes consistently improve performance. $\kappa=10$ achieves the best results with gains of 0.3--1.2\% over $\kappa=1$, particularly notable at higher IPC values (50, 100). For instance, at IPC 50, $\kappa=10$ reaches 85.6\% compared to 84.4\% for $\kappa=1$. We use $\kappa=3$ as default for the best accuracy-efficiency trade-off.

\subsection{Component Ablation Study}
\label{supp:sec:component_ablation}

We conduct a systematic ablation study to analyze the contribution of each component in our method. Starting from the DiT baseline, we progressively add: (1) learnability-guided generation, (2) deviation guidance for diversity, and (3) learnability-based ranking for sample selection. \cref{supp:tab:component_ablation} shows results across IPC 20, 30, 40, and 50.

Each component provides consistent improvements across all IPC settings. Learnability guidance alone yields 4.7--7.0\% gains over the DiT baseline, demonstrating that conditioning generation on model learning state produces more informative samples. Adding deviation guidance further improves performance by 1.9--3.3\%, showing that explicitly encouraging diversity helps avoid redundancy. Finally, learnability ranking contributes an additional 0.3--1.2\% by selecting the most learnable samples from a larger candidate pool. The full method achieves cumulative improvements of 8.1--10.0\% over DiT baseline, with the largest gains at higher IPC values where redundancy is more problematic. Notably, the improvements are additive and consistent, validating that each component addresses a distinct aspect of high-quality dataset distillation.

\begin{table}[t]
\centering
\caption{Component ablation study on ImageNette. We progressively add components starting from DiT baseline. Mean$\pm$std accuracy (\%) across 3 runs with ResNet-AP-10.}
\label{supp:tab:component_ablation}
\small
\setlength{\tabcolsep}{2.5pt}   
\renewcommand{\arraystretch}{0.92} 
\resizebox{\linewidth}{!}{%
\begin{tabular}{lcccc}
\toprule
\textbf{Method}& \textbf{IPC 20} & \textbf{IPC 30} & \textbf{IPC 40} & \textbf{IPC 50} \\
\midrule
DiT (Baseline) & 68.2$\pm$1.1 & 73.0$\pm$1.7 & 75.2$\pm$1.8  & 75.6$\pm$1.1  \\
\;+ Learnability Guidance & 75.2$\pm$0.7 & 77.8$\pm$ 0.2 & 79.9$\pm$0.7 & 82.5$\pm$0.3 \\
\;+ Deviation Guidance &  77.5 $\pm$ 0.5 & 79.9$\pm$ 0.3 & 83.2$\pm$0.8 & 84.4$\pm$1.6 \\
\;+ Learnability Ranking (Ours) & \textbf{77.8$\pm$ 0.4} & \textbf{81.1$\pm$1.5} & \textbf{83.9$\pm$ 0.8} & \textbf{85.6$\pm$0.5} \\
\midrule
\multicolumn{5}{l}{\textit{Improvements over baseline:}} \\
\;Learnability Guidance  &  +7.0  & +4.8 & +4.7 & +6.9 \\
\;+ Deviation Guidance  &  +9.3  & +6.9 & +8.0 & +8.8 \\
\;+ Ranking (Ours) & +9.6  &  +8.1 &  +8.7 & +10.0 \\
\bottomrule
\end{tabular}
} 
\end{table}

\section{Incremental Learning Details}
\label{supp:sec:incremental_algorithm}
Our method employs an incremental training strategy where synthetic data is progressively generated and added to the training set based on the current model's learning state. For a target dataset size of 100 IPC, we divide the process into $K=10$ increments, each adding 10 IPC.

\subsection{Algorithm Overview}
At each increment $i \in \{1, 2, ..., K\}$, we: (1) train learner model $\theta_i$ on current synthetic dataset $\mathcal{D}_{i-1}$, (2) generate candidate pool using diffusion model with learnability-guided selection, (3) select top samples to form increment $\mathcal{I}_i$, and (4) update dataset $\mathcal{D}_i = \mathcal{D}_{i-1} \cup \mathcal{I}_i$.

The complete procedure is detailed in \cref{alg:learnability_guided}, which presents the full incremental distillation framework. The key innovation lies in our learnability-guided diffusion sampling (\cref{alg:guided_sampling}), which conditions the generation process on both the current learner state $\theta_{i-1}$ and a reference model $\theta^*$ to synthesize samples that are informative yet semantically valid.

\paragraph{Incremental Generation and Selection.} Unlike prior methods that generate all samples independently, our approach merges candidate generation with selection to maintain diversity (\cref{alg:learnability_guided}, lines 6-16). For each class $c$ and each of the $N$ samples needed, we:
\begin{enumerate}[itemsep=0pt,topsep=2pt]
    \item Generate $\kappa$ candidate samples using learnability-guided diffusion
    \item Score candidates using the learnability criterion: $S(\mathbf{x}, c) = \mathcal{L}(\theta_{i-1}, \mathbf{x}, c) - \omega \cdot \mathcal{L}(\theta^*, \mathbf{x}, c)$
    \item Select the highest-scoring sample and add it to both the increment $\mathcal{I}_i$ and memory $\mathcal{M}_c$
\end{enumerate}
This greedy selection ensures that each newly selected sample benefits from deviation guidance against all previously selected samples within the same increment, maximizing intra-increment diversity.

\begin{algorithm}[t]
\small
\caption{LGD Distillation}
\label{alg:learnability_guided}
\KwInput{
    Pretrained diffusion model $G_{\phi}$; reference model $\theta^\star$;\\
    seed dataset $\mathcal{D}_1$; number of classes $C$; images per class $N$;\\
    number of increments $K$; over-generation factor $\kappa$;\\
    guidance strength $\lambda$; reference weight $\omega$; deviation strength $\gamma$.
}
\KwOutput{Distilled dataset $\mathcal{D}_S = \bigcup_{i=1}^{K} \mathcal{I}_i$.}

\BlankLine
\tcp{Initialize learner and per-class memory}
$\theta_1 \leftarrow \textsc{Train}(\mathcal{D}_1)$\;
\For{$c \in \{1,\dots,C\}$}{
    $\mathcal{M}_c \leftarrow \{(\mathbf{x},y) \in \mathcal{D}_1 \mid y = c\}$\;
}

\BlankLine
\tcp{Incremental distillation}
\For{$i \leftarrow 2$ \KwTo $K$}{
    $\mathcal{I}_i \leftarrow \emptyset$\;

    \For{$c \leftarrow 1$ \KwTo $C$}{
        \For{$n \leftarrow 1$ \KwTo $N$}{
            $\mathcal{C}_c^{n} \leftarrow \emptyset$\;

            \tcp{Generate $\kappa$ candidates for position $n$ in class $c$}
            \For{$k \leftarrow 1$ \KwTo $\kappa$}{
                sample $\mathbf{x}_{c,n,k} \sim
                G_{\phi}(\cdot \mid \theta_{i-1}, \theta^\star, c, \lambda, \omega, \gamma, \mathcal{M}_c)$\;
                $\mathcal{C}_c^{n} \leftarrow \mathcal{C}_c^{n} \cup \{(\mathbf{x}_{c,n,k}, c)\}$\;
            }

            \tcp{Score candidates by learnability}
            \ForEach{$(\mathbf{x}, c) \in \mathcal{C}_c^{n}$}{
                $S(\mathbf{x}, c) \leftarrow
                \mathcal{L}(\theta_{i-1}, \mathbf{x}, c)
                - \omega\, \mathcal{L}(\theta^\star, \mathbf{x}, c)$\;
            }
            $(\mathbf{x}_c^\star, c) \leftarrow
            \arg\max_{(\mathbf{x},c) \in \mathcal{C}_c^{n}} S(\mathbf{x}, c)$\;

            \tcp{Add selected sample to increment and memory}
            $\mathcal{I}_i \leftarrow \mathcal{I}_i \cup \{(\mathbf{x}_c^\star, c)\}$\;
            $\mathcal{M}_c \leftarrow \mathcal{M}_c \cup \{(\mathbf{x}_c^\star, c)\}$\;
        }
    }

    \BlankLine
    \tcp{Update cumulative dataset and retrain learner}
    $\mathcal{D}_i \leftarrow \mathcal{D}_{i-1} \cup \mathcal{I}_i$\;
    $\theta_i \leftarrow \textsc{Train}(\mathcal{D}_i)$\;
}

\BlankLine
$\mathcal{D}_S \leftarrow \mathcal{D}_K$\;

\end{algorithm}

\begin{algorithm}[t]
\small
\caption{LGD Sampling}
\label{alg:guided_sampling}

\KwInput{
Current learner $\theta_{i-1}$; reference model $\theta^\star$; class $c$;\\
guidance strength $\lambda$; reference weight $\omega$; deviation strength $\gamma$;\\
diffusion model $G_\phi$; number of timesteps $T$; per-class memory $\mathcal{M}_c$.
}
\KwOutput{Generated sample $\mathbf{x}_0$.}

\BlankLine
\tcp{Initialize from Gaussian noise}
Sample $\mathbf{x}_T \sim \mathcal{N}(\mathbf{0}, \mathbf{I})$\;

\For{$t \leftarrow T$ \KwTo $1$}{
    
    \tcp{Standard diffusion denoising prediction}
    $\boldsymbol{\epsilon}_\phi(\mathbf{x}_t, t, c)
        \leftarrow G_\phi(\text{predict-noise} \mid \mathbf{x}_t, t, c)$\;

    \BlankLine
    \If{$t \in [10,45]$}{
        
        \tcp{Compute learnability score}
        $S(\mathbf{x}_t,c) \leftarrow
            \mathcal{L}(\theta_{i-1},\mathbf{x}_t,c)
            - \omega\, \mathcal{L}(\theta^\star,\mathbf{x}_t,c)$\;

        \tcp{Timestep-dependent scaling (normalizes gradient magnitudes)}
        $\rho_t \leftarrow
        \dfrac{
            \sqrt{1 - \bar{\alpha}_t} \cdot
            \|\boldsymbol{\epsilon}_\phi(\mathbf{x}_t,t,c)\|
        }{
            \|\nabla_{\mathbf{x}_t} S(\mathbf{x}_t,c)\|
        }$\;

        \tcp{Learnability guidance}
        $\tilde{\boldsymbol{\epsilon}}_\phi(\mathbf{x}_t,t,c)
            \leftarrow
            \boldsymbol{\epsilon}_\phi(\mathbf{x}_t,t,c)
            + \lambda\, \rho_t \nabla_{\mathbf{x}_t} S(\mathbf{x}_t,c)$\;

        \tcp{Deviation guidance for diversity}
        $\tilde{\mathbf{x}}^\star
            \leftarrow \argmin_{\tilde{\mathbf{x}} \in \mathcal{M}_c}
                \|\mathbf{x}_t - \tilde{\mathbf{x}}\|$\;
        $\mathcal{G}_D(\mathbf{x}_t)
            \leftarrow
            \dfrac{\mathbf{x}_t \cdot \tilde{\mathbf{x}}^\star}{
                \|\mathbf{x}_t\|\, \|\tilde{\mathbf{x}}^\star\|
            }$\;

        $\tilde{\boldsymbol{\epsilon}}_\phi(\mathbf{x}_t,t,c)
            \leftarrow
            \tilde{\boldsymbol{\epsilon}}_\phi(\mathbf{x}_t,t,c)
            - \gamma\, \nabla_{\mathbf{x}_t}\mathcal{G}_D(\mathbf{x}_t)$\;
    }
    \Else{
        $\tilde{\boldsymbol{\epsilon}}_\phi(\mathbf{x}_t,t,c)
        \leftarrow
        \boldsymbol{\epsilon}_\phi(\mathbf{x}_t,t,c)$\;
    }

    \BlankLine
    \tcp{Reverse diffusion update}
    $\boldsymbol{\mu}_\phi(\mathbf{x}_t)
    \leftarrow
    \dfrac{1}{\sqrt{1 - \beta_t}}
    \left(
        \mathbf{x}_t -
        \dfrac{\beta_t}{\sqrt{1 - \bar{\alpha}_t}}
        \tilde{\boldsymbol{\epsilon}}_\phi(\mathbf{x}_t,t,c)
    \right)$\;

    Sample $\mathbf{z} \sim \mathcal{N}(\mathbf{0},\mathbf{I})$\;
    $\mathbf{x}_{t-1} \leftarrow \boldsymbol{\mu}_\phi(\mathbf{x}_t) + \sigma_t \mathbf{z}$\;
}

\Return{$\mathbf{x}_0$}\;

\end{algorithm}

\paragraph{Reference Model.} For learnability scoring, we trained a reference model on the full dataset without MixCut for 300 epochs on the target dataset ImageNette or ImageWoof. For ImageNet-1K we used a pretrained ResNet-18 network provided by PyTorch.

\paragraph{Learner Training.} We train the learner from scratch using the initial increment $\mathcal{D}_1$ (always at IPC 10). Training proceeds until the validation loss plateaus, using a patience of 300. We adopt the AdamW optimizer with a learning rate of 0.001, $\beta_0 = 0.9$, $\beta_1 = 0.999$, and a weight decay of $0.01$.
After convergence, we further fine-tune the model using a cosine-decay schedule with a minimum learning-rate ratio of $0.01$ for 300 epochs, which effectively ``squeezes'' the training signal available from the current distilled data. In the case of ImageNet-1k we train the model 300 epochs using the setup described in Soft-Label Protocol in \cref{supp:sec:implementation}

\paragraph{Learner Model for Synthesis.}
Because the learner is trained on a small distilled dataset, its instantaneous gradient updates can be noisy and may not reliably reflect the underlying training signal. To obtain a more stable representation of the model's evolving knowledge, we maintain an exponential moving average (EMA) of the learner throughout training. This EMA model is then used to compute learnability guidance, ensuring smoother and more reliable gradient estimates for synthesis.

\paragraph{Memory Management.} The memory sets $\mathcal{M}_c$ play a crucial role in deviation guidance. Initialized from the seed dataset $\mathcal{D}_1$ (\cref{alg:learnability_guided}, line 3), they are incrementally updated as new samples are selected (line 16). During diffusion sampling, the memory ensures generated samples push away from all previously synthesized examples of the same class, preventing redundancy and promoting diversity.

\subsection{Incremental Evaluation During Synthesis}
\label{supp:sec:incremental_eval}

During the learnability-guided distillation process (\cref{alg:learnability_guided}), the learner model $\theta_i$ is trained incrementally on the growing distilled dataset $\mathcal{D}_i$ at each stage (line 18). We can evaluate this learner model on the test set and report its top-1 accuracy at the corresponding IPC. This provides insight into the quality of the distilled dataset as it is being constructed, showing how model performance evolves with each added increment. For example, after generating increment $\mathcal{I}_2$ and training $\theta_2$ on $\mathcal{D}_2 = \mathcal{D}_1 \cup \mathcal{I}_2$ (20 IPC total), we can evaluate $\theta_2$ and report accuracy at IPC 20 before proceeding to generate increment $\mathcal{I}_3$.

This incremental evaluation is distinct from the standard static evaluation protocol used in Tables 1-3 of the main paper, where a fresh model is trained from scratch on the complete final distilled dataset $\mathcal{D}_K$. The incremental evaluation uses the same model that guides the synthesis process—the learner that has been progressively trained on $\mathcal{D}_1, \mathcal{D}_2, ..., \mathcal{D}_i$—whereas static evaluation trains a new model only on $\mathcal{D}_K$. The incremental accuracy curve $\{(i \cdot N, \text{Acc}(\theta_i))\}_{i=1}^{K}$ shown in \cref{fig:val} and \cref{tab:progressive} directly reflects the learning dynamics during distillation, revealing sustained marginal gains as each curriculum-aligned increment is added.


\section{Implementation Details}
\label{supp:sec:implementation}

Here, we provide comprehensive implementation details for reproducibility of our experiments, including network architectures, hyperparameter settings. All experiments are conducted in a single NVIDIA H100 (80GB) GPU.

\subsection{Diffusion Model Configuration}
For diffusion model pre-training, we adopt DiT-XL/2 256$\times$256 following the settings ~\cite{gu2024efficient, chen2025igd}. Generated images are at $256\times256$ resolution. We use 50 DDPM sampling steps with classifier-free guidance scale 4.0. Generated images are clipped to $[-1, 1]$ and saved as PNG format.

\subsection{Evaluation Settings}

We outline the evaluation settings used to assess the performance of the distilled datasets.

\subsubsection{Network Architectures}
\label{supp:sec:architectures}

We conduct experiments on three commonly adopted network architectures in the area of dataset distillation:

\begin{enumerate}
    \item \textbf{ConvNet-6} is a 6-layer convolutional network. The network contains 128 feature channels in each layer, and instance normalization is adopted.
    \item \textbf{ResNetAP-10} is a 10-layer ResNet, where the strided convolution is replaced by average pooling for downsampling.
    \item \textbf{ResNet-18} is a 18-layer ResNet with instance normalization.
\end{enumerate}


\begin{table}[t]
\centering
\caption{Hyperparameter settings for data synthesis and learnability scoring.}
\label{supp:tab:hyperparams_synthesis}
\small
\setlength{\tabcolsep}{4pt}
\begin{tabular}{@{}lll@{}}
\toprule
\textbf{Config} & \textbf{Value} & \textbf{Explanation} \\
\midrule
\multicolumn{3}{c}{\textit{(a) Diffusion Synthesis}} \\
\midrule
Sampler & DDPM & -- \\
Sampling Steps & 50 & -- \\
CFG Scale & 4.0 & Classifier-free guidance \\
\midrule
\multicolumn{3}{c}{\textit{(b) Learnability Guidance}} \\
\midrule
Learner Arch. & ResNet-10-AP & -- \\
Reference Arch. & ResNet-10-AP & Same as learner \\
Ref.\ Weight $\omega$ & 0.5 & Eq.~7 in main paper \\
Guidance $\lambda$ & 15 & Eq.~8 in main paper \\
Deviation $\gamma$ & 50 & Diversity control \\
Learnability Batch $K$ & 3 & Scoring batch size \\
\midrule
\multicolumn{3}{c}{\textit{(c) Incremental Configuration}} \\
\midrule
Samples/Increment & 10 & samples/class \\
Seed Init. & IGD & Initial 10 IPC \\
\bottomrule
\end{tabular}
\end{table}

\paragraph{Hard-Label Protocol.}
For validation training, we follow the protocol established in~\cite{chen2025igd}. We use AdamW with a learning rate of 0.001, momentum parameters $\beta_0 = 0.9$ and $\beta_1 = 0.999$, and a weight decay of 0.01. The number of training epochs for each IPC setting is 2000, 1500, 1500, 1500, 1500, and 1000 for IPC 10, 20, 30, 40, 50, and 100, respectively. We apply learning rate decays at the 2/3 and 5/6 points of training, using a decay factor (gamma) of 0.2.

\paragraph{Soft-Label Protocol.} We follow the evaluation protocol used by~\cite{gu2024efficient, sun2024rded} for ImageNet-1k evaluation. We train ResNet-18 for 300 epochs using AdamW optimizer with learning rate 0.001, weight decay 0.01, and parameters $\beta_1 = 0.9$ and $\beta_2 = 0.999$. We employ smoothing LR schedule with rand augment and RandomHorizontalFlip.

\section{Qualitative Analysis}
\label{supp:sec:qualitative}

\begin{figure*}[tbh]
    \centering
    \includegraphics[page=1,width=\linewidth]{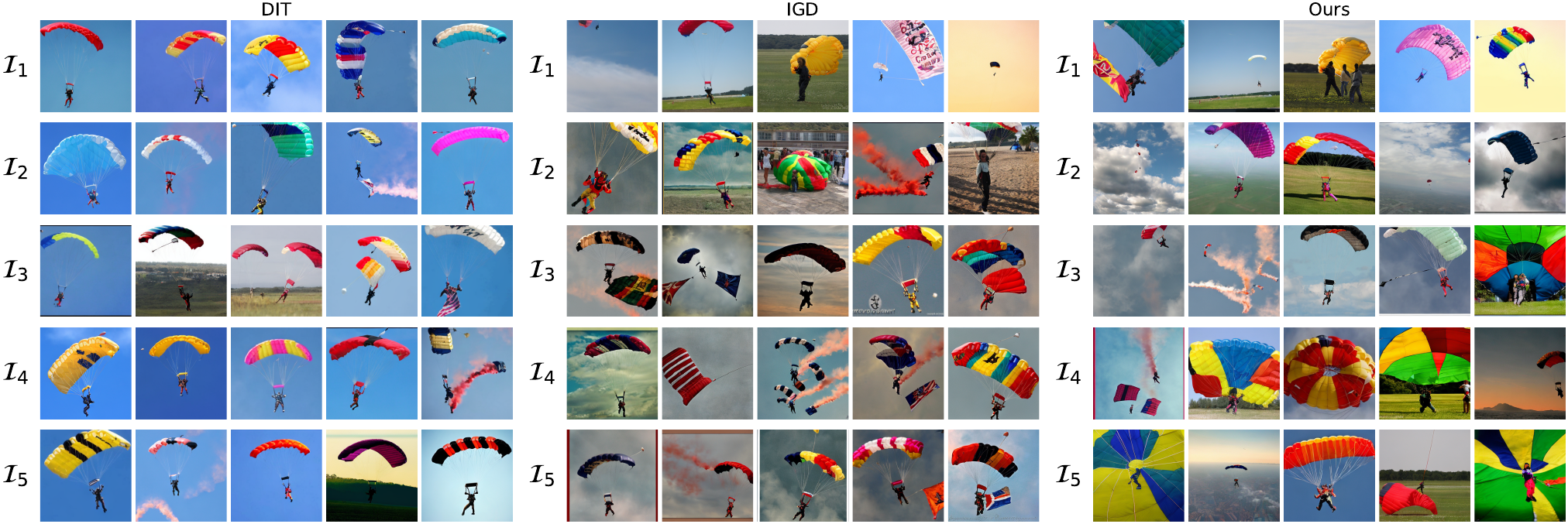}
\caption{\textbf{Visual diversity in incrementally distilled datasets.}
Samples from increments
$\increment_1  -\increment_5$ (50 IPC total) of the Parachute class.}
\label{fig:parachute_diversity}
\end{figure*}

\begin{figure*}[tbh]
    \centering
    \includegraphics[page=1,width=\linewidth]{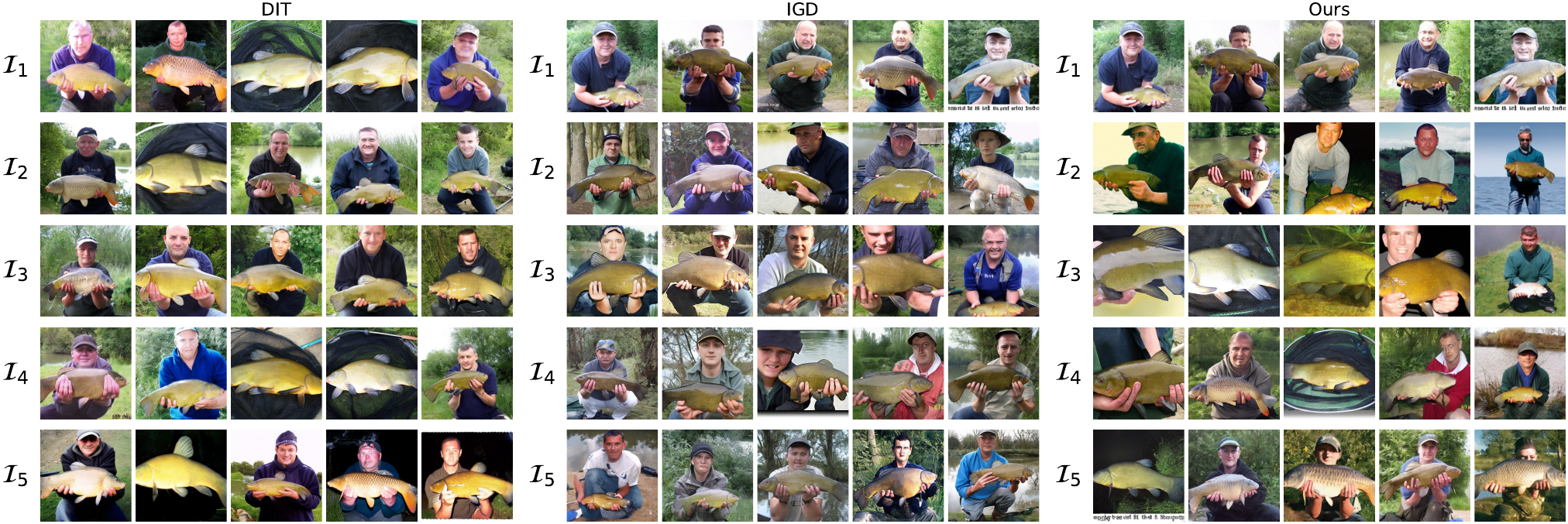}
\caption{\textbf{Visual diversity in incrementally distilled datasets.}
Samples from increments
$\increment_1  -\increment_5$ (50 IPC total) of the Tench class.}
\label{fig:tench_diversity}
\end{figure*}

\begin{figure*}[tbh]
    \centering
    \includegraphics[page=1,width=\linewidth]{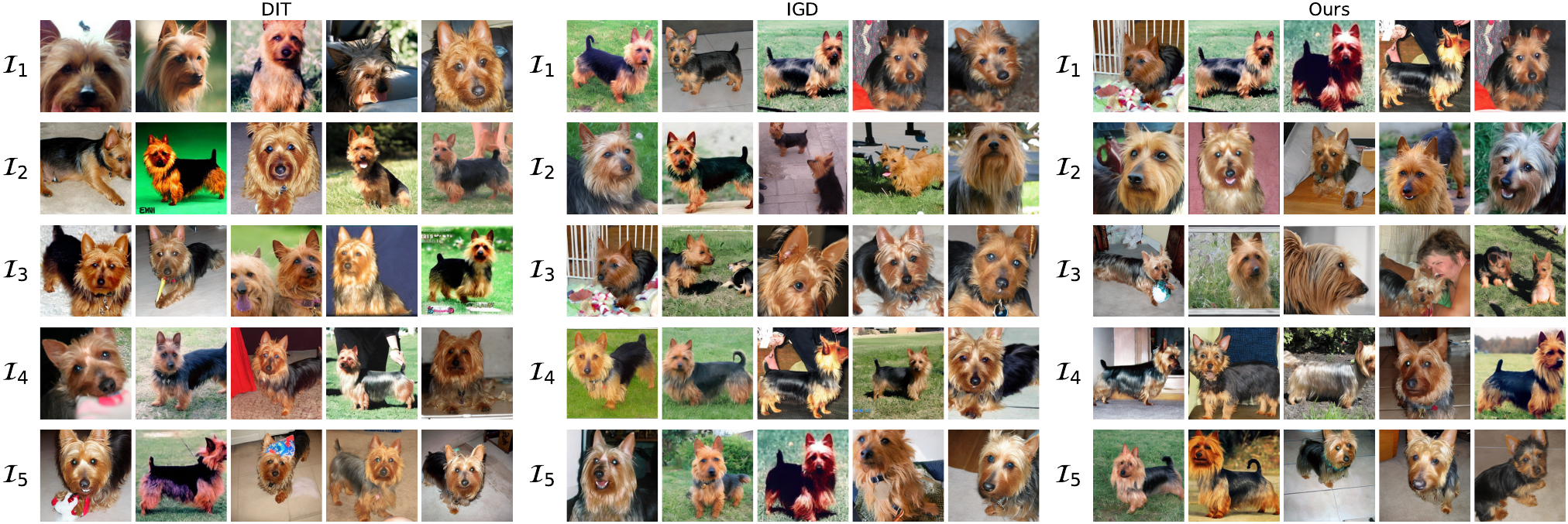}
\caption{\textbf{Visual diversity in incrementally distilled datasets.}
Samples from increments
$\increment_1  -\increment_5$ (50 IPC total) of the Australian terrier class.}
\label{fig:terrier_diversity}
\end{figure*}

\begin{figure*}[tbh]
    \centering
    \includegraphics[page=1,width=\linewidth]{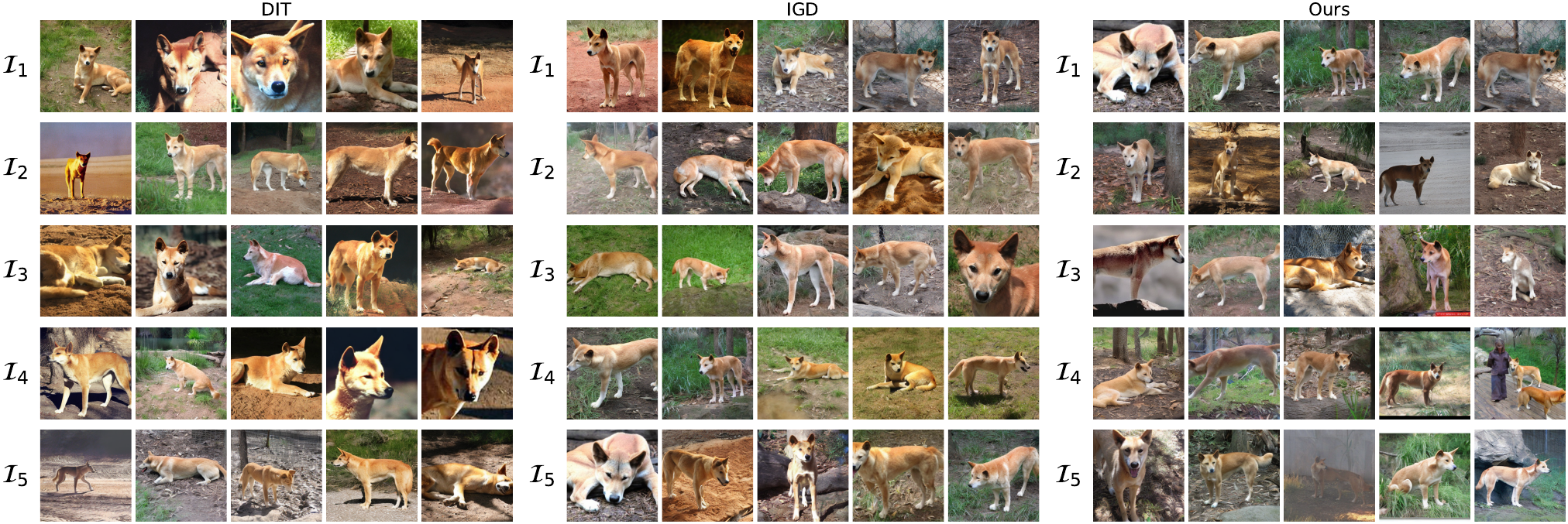}
\caption{\textbf{Visual diversity in incrementally distilled datasets.}
Samples from increments
$\increment_1  -\increment_5$ (50 IPC total) of the Dingo class.}
\label{fig:dingo_diversity}
\end{figure*}

We provide qualitative visualizations comparing samples generated by our learnability-guided approach against baseline methods. \cref{fig:parachute_diversity,fig:tench_diversity,fig:terrier_diversity,fig:dingo_diversity} show representative samples from ImageNette and ImageWoof datasets across different methods.

These visualizations reveal that our method generates samples with greater diversity and semantic richness compared to baseline approaches. While vanilla diffusion models and IGD tend to produce visually similar samples within each increment, our learnability-guided synthesis yields distinct variations that capture complementary visual features, consistent with our quantitative complementarity analysis in \cref{supp:sec:complementarity}.

\end{document}